\documentclass[]{fairmeta}

\title{\resizebox{\linewidth}{!}{Digitizing Touch with an Artificial Multimodal Fingertip}}

\usepackage{epsfig}
\usepackage{epstopdf}
\usepackage{nameref}
\usepackage{hyperref}
\usepackage{gensymb}
\usepackage{mathtools}
\usepackage{caption}
\usepackage{subcaption}
\usepackage{wrapfig}
\usepackage{siunitx}
\usepackage{float}
\usepackage{floatflt}
\usepackage{url}
\usepackage{bm}
\usepackage{multibib}
\usepackage[version=4]{mhchem}
\usepackage{graphicx}
\usepackage{multirow}
\usepackage{amsmath,amssymb,amsfonts}
\usepackage{amsthm}
\usepackage{mathrsfs}
\usepackage[title]{appendix}
\usepackage{xcolor}
\usepackage{textcomp}
\usepackage{manyfoot}
\usepackage{booktabs}
\usepackage{algorithm}
\usepackage{algorithmicx}
\usepackage{algpseudocode}
\usepackage{listings}

\newcommand{\sensor}[0]{Digit 360}

\author[1]{Mike Lambeta}
\author[1]{Tingfan Wu}
\author[1]{Ali Seng\"ul}
\author[1]{Victoria Rose Most}
\author[1]{Nolan Black}
\author[1]{Kevin Sawyer}
\author[1]{Romeo Mercado}
\author[1,4]{Haozhi Qi}
\author[1]{Alexander Sohn}
\author[1]{Byron Taylor}
\author[1]{Norb Tydingco}
\author[1]{Gregg Kammerer}
\author[1]{Dave Stroud}
\author[1]{Jake Khatha}
\author[1]{Kurt Jenkins}
\author[1]{Kyle Most}
\author[1]{Neal Stein}
\author[1]{Ricardo Chavira}
\author[1]{Thomas Craven-Bartle}
\author[1]{Eric Sanchez}
\author[1]{Yitian Ding}
\author[1]{Jitendra Malik}
\author[2,3]{Roberto Calandra}

\affiliation[1]{FAIR at Meta}
\affiliation[2]{Learning, Adaptive Systems, and Robotics (LASR) Lab, TU Dresden}
\affiliation[3]{The Centre for Tactile Internet with Human-in-the-Loop (CeTI)}
\affiliation[4]{University of California, Berkeley}

\abstract{Touch is a crucial sensing modality that provides rich information about object properties and interactions with the physical environment. Humans and robots both benefit from using touch to perceive and interact with the surrounding environment \citep{Johansson2009Coding,Li2020review,Calandra2017Feeling}. However, no existing systems provide rich, multi-modal digital touch-sensing capabilities through a hemispherical compliant embodiment. Here, we describe several conceptual and technological innovations to improve the digitization of touch. These advances are embodied in an artificial finger-shaped sensor with advanced sensing capabilities. Significantly, this fingertip contains high-resolution sensors ($\approx$8.3 million taxels) that respond to omnidirectional touch, capture multi-modal signals, and use on-device artificial intelligence to process the data in real time. Evaluations show that the artificial fingertip can resolve spatial features as small as 7 um, sense normal and shear forces with a resolution of 1.01 mN and 1.27 mN, respectively, perceive vibrations up to 10 kHz, sense heat, and even sense odor. Furthermore, it embeds an on-device AI neural network accelerator that acts as a peripheral nervous system on a robot and mimics the reflex arc found in humans. These results demonstrate the possibility of digitizing touch with superhuman performance. The implications are profound, and we anticipate potential applications in robotics (industrial, medical, agricultural, and consumer-level), virtual reality and telepresence, prosthetics, and e-commerce. Toward digitizing touch at scale, we open-source a modular platform to facilitate future research on the nature of touch.}

\date{October 31, 2024}
\correspondence{Roberto Calandra (\email{rcalandra@lasr.org}) and Mike Lambeta (\email{lambetam@meta.com})}

\metadata[Code]{\url{https://github.com/facebookresearch/digit360}}
\metadata[Website]{\url{https://digit.ml}}

\begin{document}

\maketitle

\section{Introduction}\label{sec1_introduction}

\begin{figure}[h!]   
    \centering
    \includegraphics[width=1\linewidth]{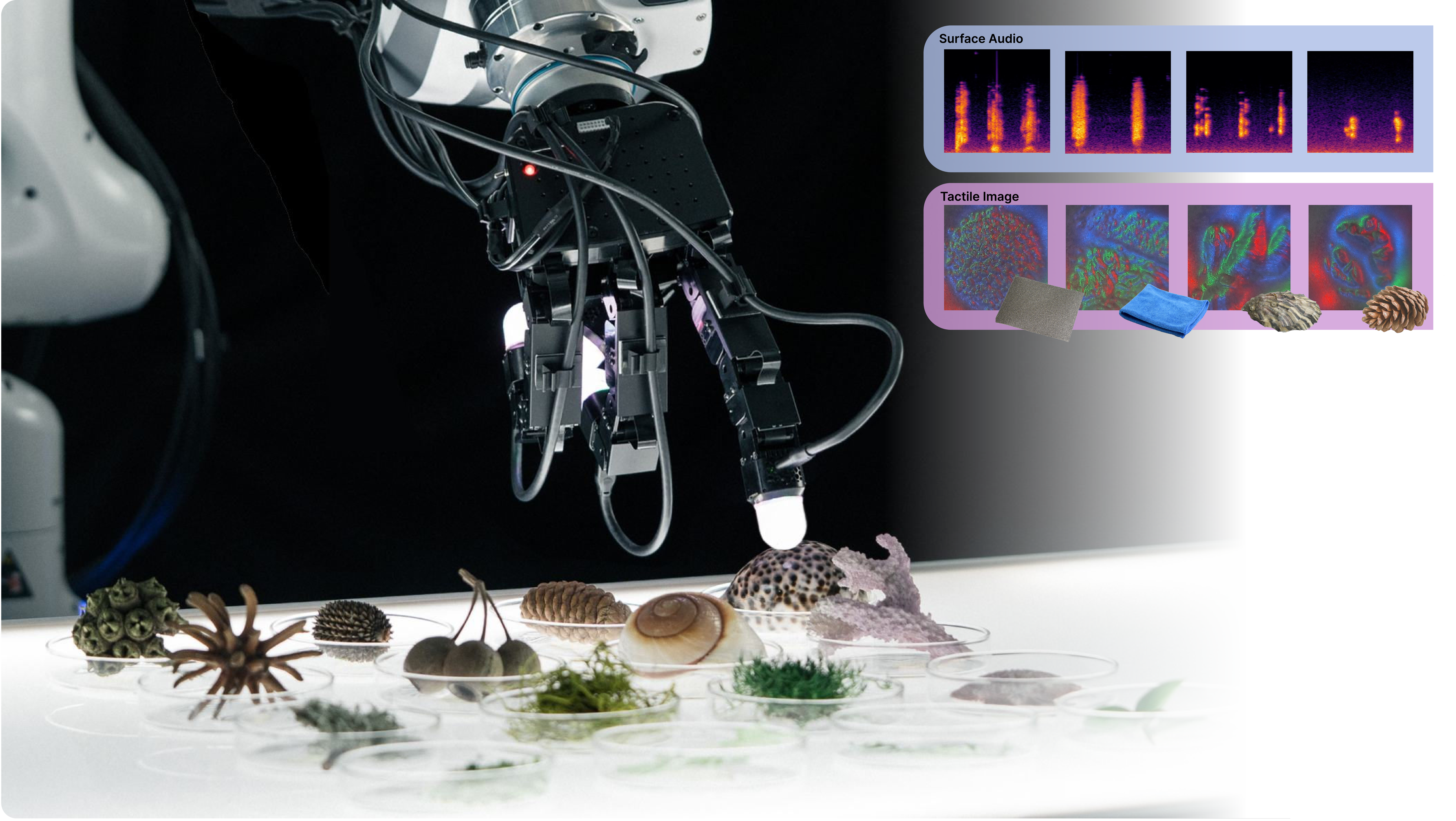}
    \caption{Digitizing touch with the artificial multimodal fingertip.}
    \label{fig:fig_teaser}
\end{figure}

While artificial intelligence (AI) has the ability to understand and manipulate language, is AI able to palpably distinguish between a rough surface and a smooth surface? A raw egg and a hard-boiled egg? A soft, pliant surface and a firm, soft or rigid surface? The proverbial full glass and a half-empty glass? 
In short, how can AI bridge the gap between the digital and physical worlds? Of all our senses, touch arguably is the most critical in how we interact with the world~\citep{Johansson2009Coding} and how we explore the world~\citep{lederman1987hand}. 
It enables us to measure forces and recognize object properties – shape, weight, density, textures, friction, and elasticity. It also plays an important role both in social relationships~\citep{Dunbar2010social} and in cognitive development~\citep{Ardiel2010importance}. 
Until now, no solutions have emerged for digitizing touch with the same rich sensorial spectrum that is inherent to the human experience~\citep{klatzky1992stages}.
Toward the advancement of robotic in-hand manipulation, we strive to mimic familiar features of the human hand: Fingers. 
We postulate that the next generation of touch digitization in robotics will enable intelligent systems to discern significantly higher levels of physical information during environmental interaction. 

Digitizing touch depends on two fundamental features: one temporal in nature and one spatial. 
Temporal features process information from signals of time variation, whereas spatial features process a discrete multidimensional array of geometrical signals. 
We combine these methods within a unified platform to significantly improve the capabilities of touch digitization: a modular, finger-shaped, multi-modal tactile sensor with on-device artificial intelligence capabilities and superhuman performance. 
Previous efforts in the field have involved subsets of sensing modalities optimized for cost~\citep{Lambeta2020DIGIT}, fingertip geometries iterations~\citep{Padmanabha2020OmniTact,Romero2020Soft,Gomes2020,Choi2020Dexterous,Sun2022soft}, or for maximizing a given design metric. 
However, those choices led to limitations in performance. 
The primary sensing modality of vision-based tactile sensors captures the geometry of the object being touched. 
From that geometric data, it is possible to reconstruct normal and shear forces. 
However, that particular modality falls short of the rich, multi-modal nature of human skin, which uses many different types of receptors~\citep{Johnson2001roles,Dahiya2009Tactile,Handler2021mechanosensory} (such as mechanoreceptors, thermoreceptors, and nociceptors). 

\begin{figure}[htbp]
    \centering
    \includegraphics[width=\textwidth]{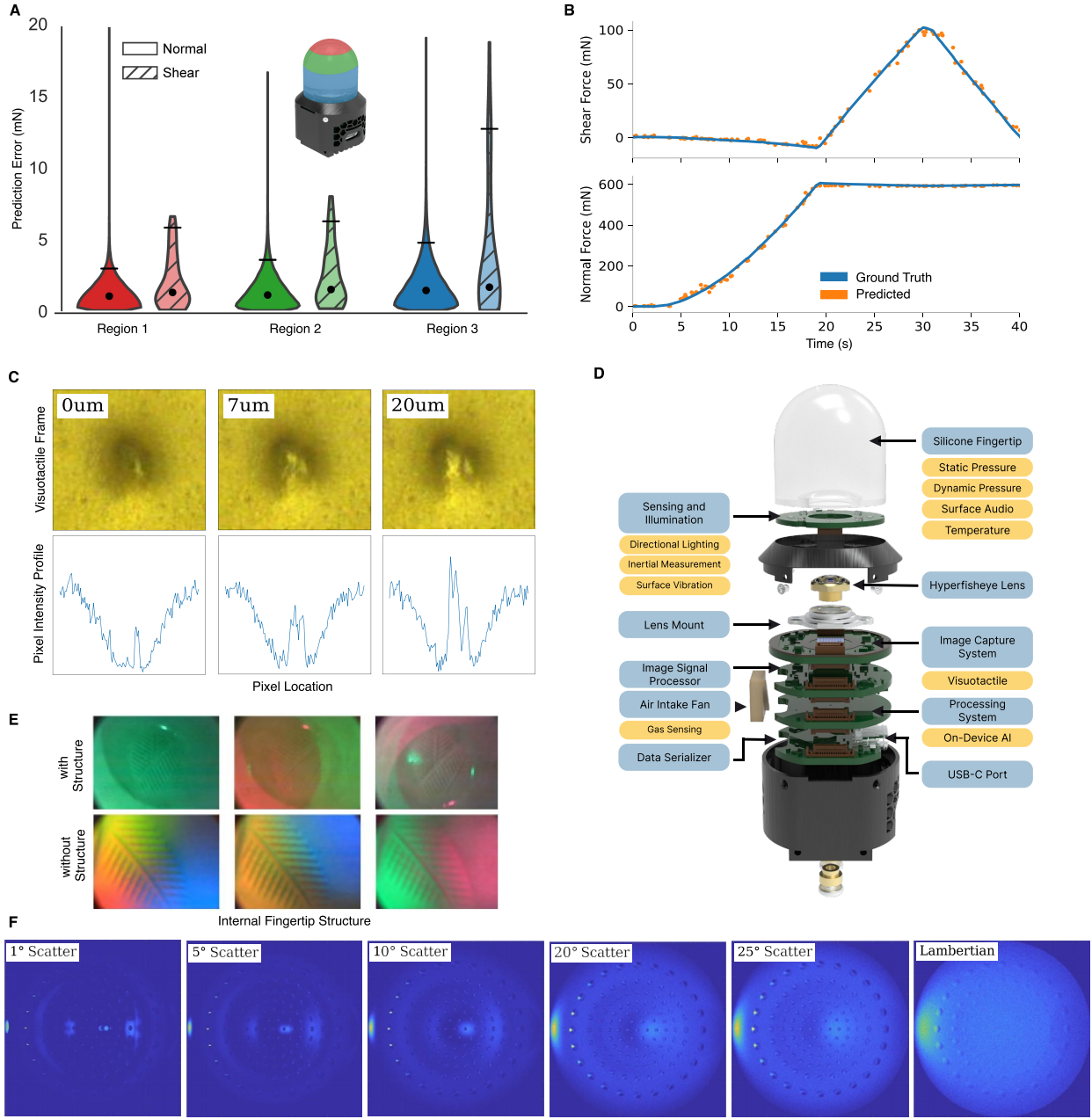}
    \vspace{-9em}
    \caption{
    \textbf{a)} We evaluated normal and shear forces in three separate regions of the sensor, from the tip toward the side. The dots and error bars show the median and 95-percentile of the error, respectively. The median error from the deep-learning model for the three regions is 1.01 mN, 1.09 mN, 1.41 mN for normal forces, and 1.27 mN, 1.48 mN, 1.64 mN for shear forces. \textbf{b)} We train a deep learning model to predict normal and shear forces from visuo-tactile image output. Normal and shear forces are predicted with a median error of 1.01 mN and 1.27 mN, respectively. Compared to alternative methods, predicting shear force requires the use of markers; however, with increased spatial resolution, far more features are extracted from the visuo-tactile image, which aids in shear force prediction. \textbf{c)} We evaluate spatial resolution by using a dual-pronged microindenter depressed into the artificial fingertip with varying widths. Visual validation and the inspection of the taxels’ profile intensity confirmed the ability to clearly distinguish features as small as 7 $\mu m$. \textbf{d)} We show two methods that create the artificial fingertip volume: internal structure (top row) and solid gel with immersion lens (bottom row). With an internal structure, illumination artifacts are visible, whereas, with a solid volume, the resulting image is far higher quality with fewer illumination artifacts. \textbf{e)} We simulate the effects of increasing the surface scattering along the internal reflective layer of the artificial fingertip. From left to right, machine polish of 1\degree{} to Lambertian scattering, we optimize for image contrast while constraining the background illumination uniformity.
    }
    \label{fig:fig_1}
\end{figure}

We develop and introduce a high-end modular research platform for investigating touch and digitization through novel approaches.  
Our platform, identified as Digit 360, belongs to the family of vision-based tactile sensors~\citep{Abad2020Visuotactile}: It embodies an elastomer that serves as the touch-sensing interface, with a subcutaneous camera that measures the deformation of the elastomer through structured light. 
But in addition to camera-based sensing, our platform is capable of sensing multiple tactile modes (see Figure~\ref{fig:fig_1}): Contact intensity and geometry, static and dynamic forces, surface audio textures and vibrations, heat change, fingertip velocity/acceleration/orientation, and even identification of some airborne chemical compounds. 
Further, to emulate the human reflex arc, we introduce an on-device AI neural network accelerator that provides next-generation touch-processing capabilities, enables real-time local processing to minimize reaction delays, and reduces communication bandwidth.


\section{Artificial Fingertip}
\label{sec2_ai_fingertip}

To effectively capture the nuances in touch interactions with the world, an artificial fingertip must be sensitive in both temporal and spatial domains. 
These domains are obtained as modal signals encompassed within the artificial fingertip through visual, audio, vibration, pressure, heat, and gas sensing. 
Prior vision-based touch sensors -- using off-the-shelf imaging systems -- are bound by slow visual capture rates, which reduce the amount of sequential information resulting from frame encoding time and, therefore, limit the temporal nature of non-static touch interactions encountered during manipulation. 
Increasing the temporal frequency of the visual system would not benefit without an increase in spatial resolution for dynamic movements. 
We encompass the touch digitization system into the form of an artificial fingertip with a similar geometry to a human finger. 
The surface of the fingertip encodes touch information from depressions in a reflective layer of which internal light reflections are captured by the camera. 
Ideally, a visual-based tactile system should resolve the minimum possible spatial features presented by an object interacting with the fingertip at high temporal rates. 
Our artificial fingertip, Digit 360, is designed to advance spatial, temporal, and multi-modal performance through the embodiment of a modular platform in touch digitization research.
To achieve the high spatial and temporal performance demonstrated by our artificial fingertip, we made methodological breakthroughs in five different subsystems: elastomer interface, optical system, illumination system, multi-modal sensing, and on-device AI processing.

\begin{figure}[!t]   
    \centering
        \includegraphics[width=1\linewidth]{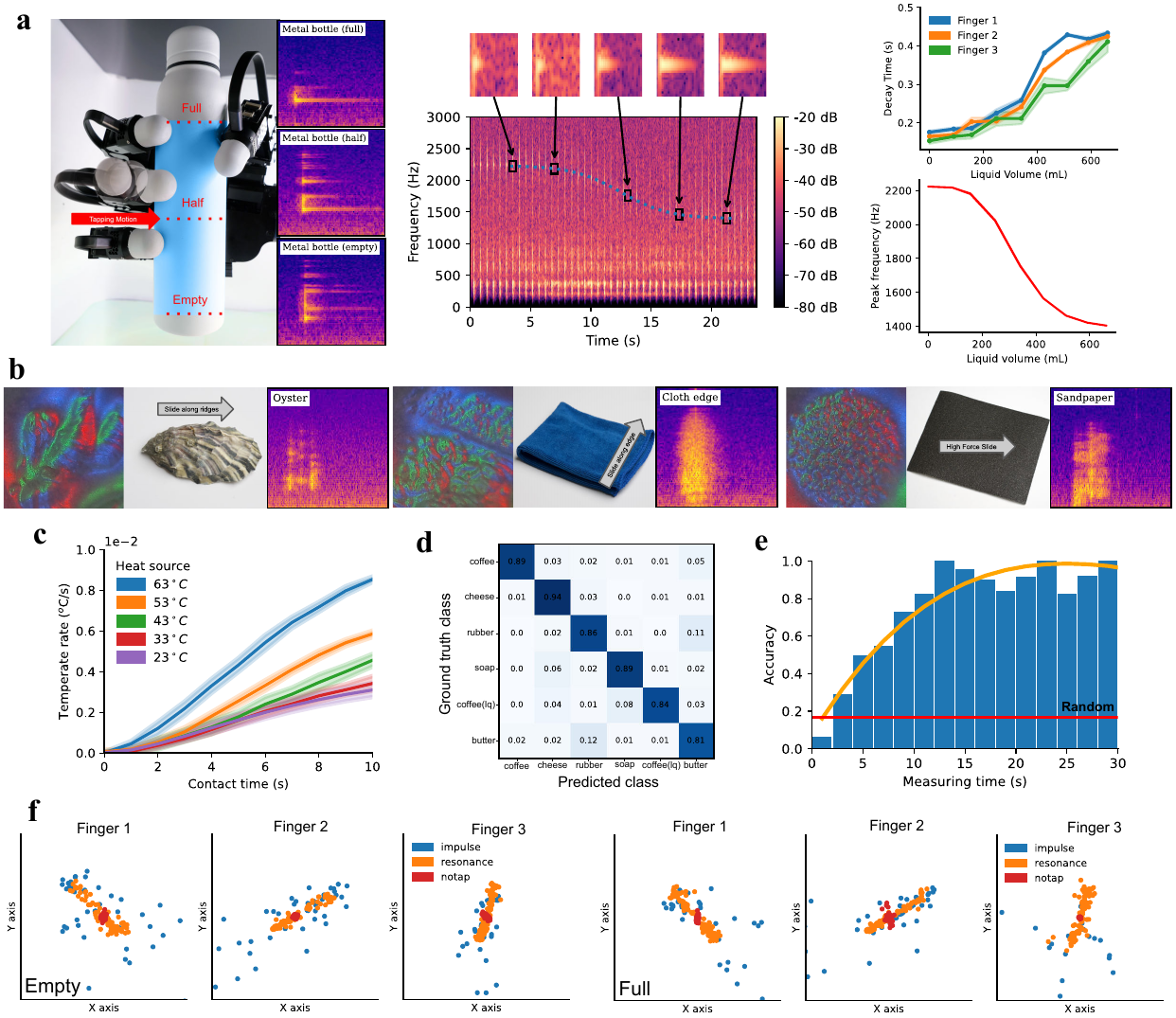}
    \caption{\textbf{a)}. We demonstrate the ability to determine the volume of water in an opaque container by tapping with one finger and recording the response on the fingers in contact with the container. We further show how this modality is deconstructed into peak frequency analysis, independent of finger position, whereas decay time depends on finger placement. \textbf{b)} Spectrogram of the surface audio textures recorded for different objects. \textbf{c)} Using a variable heat source as a control, the artificial fingertip is sensitive to heat gradients. \textbf{d)} The artificial fingertip provides object state and identification of objects through local gas sensing, achieving a 91\% accuracy.\textbf{e)} The accuracy of object classification through scent depends on the integration time as the artificial fingertip begins approaching the object. We show that 61\% accuracy is reached within 6 seconds. \textbf{f)} Localization of finger placement on an object during movement transients with empty and full volumes of liquid. We measure the effects of transients during impulse, resonance during a static hold, and a static hold with no movement.}
    \label{fig:fig_2}
\end{figure}

\subsection{Elastomer Interface}
When the reflective fingertip surface layer is subject to impression stimuli, the surface layer material properties directly relate to the spatial resolving capabilities. 
A design-of-experiment technique is developed to identify these six material parameters that affect sensor sensitivity to input stimuli: Rgel (gel fingertip radius), Tc (surface reflective coating layer thickness), Tg (surface layer thickness), h (height), Ec (coating Young’s modulus), and Eg (fingertip volume Young’s modulus). 
If Tc and Tg are too thick or exhibit low compliance, a low pass filter effect is evident on discrete object edges. 
Similarly, if the object is rich in spatial information and fractal dimension~\citep{sahli2020tactile}, the fingertip surface will resolve fewer features due to local gradients from material compliance. 
We avoid specifying any constraints on the coating thickness layer to best capture small input stimuli while maintaining a suitable parameter range for the general size of the fingertip. 
Depositing and applying this layer onto the fingertip surface could involve manual hand painting, airbrushing, or dip-coating techniques. However, while these techniques produce a touch image, they are far from optimal and result in large coating thickness and inconsistent yield from manufacturing variance. 
We solve this by developing a new chemical deposition technique for growing a silver thin film directly onto the surface of the fingertip. This technique produces coating thicknesses far smaller than previous methods and thus achieves better sensitivity.

\subsection{Optical System}
Common vision-based sensors capture input stimuli at a planar surface, use multiple cameras that are difficult to integrate and process together, or default to common off-the-shelf cameras optimized for human-centered imaging, which results in downgraded optical performance in touch. We refrain from the use of standard image-sensor features such as automatic exposure control, automatic white balance, and automatic focus which are designed for responding to changes in the natural environment as our fingertip chamber is an enclosed and controlled environment. 
We find that a new approach is required to optimize the capture of the hemispherical surface when modeling an isotropic representation of similar dimensions to a human fingertip. 
In optimizing for input stimuli from the touch interaction layer, we stipulate that our imaging system should not limit the performance within the finite element method simulation of the material properties.
Hence, we determine the optical system requirements to best suit capturing images related to tactile sensing with a CMOS pixel size of 1.1~um. 
Parameters were chosen for converging spot size to increase spatial resolution, intentionally allowing chromatic aberration, introducing shallow depth of field to allow for defocus proportional to object indentation depth, and removing anti-reflective coatings to enable capture and interpretation of reflections and scattering inside the fingertip. 
However, such parameters require a non-standard lens. Therefore, we developed a custom solid immersion hyper fisheye lens to tackle the unique environment of visuotactile sensing rather than an off-the-shelf lens catering to general-purpose imaging, thus enabling full control over lens geometry and optical parameters.

\begin{figure*}[!t]   
    \centering
    \makebox[\textwidth][c]{
        \centering
        \includegraphics[width=\linewidth]{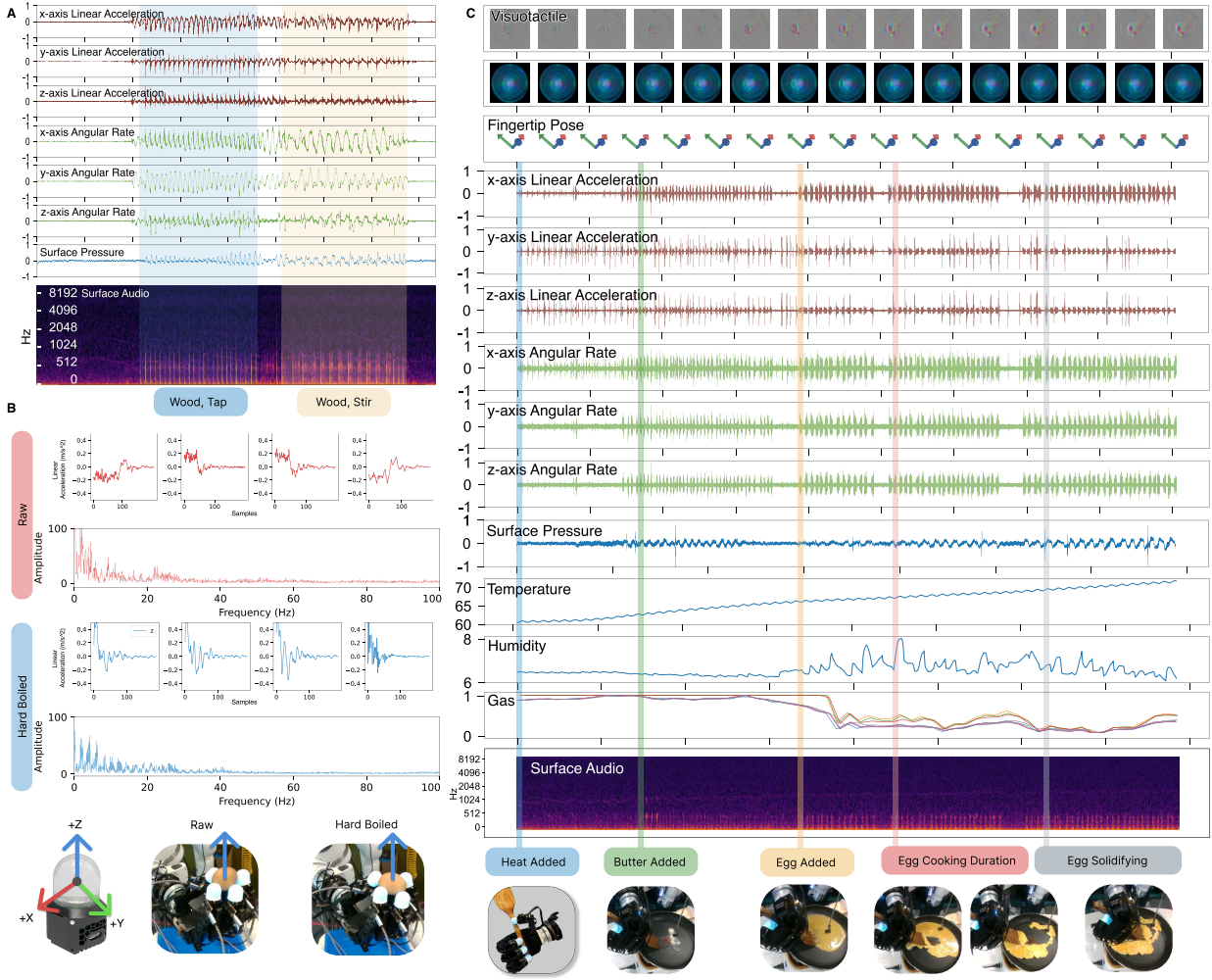}
    }
    \caption{Different actions, materials and states during touch digitization uniqueliy contirbute to excitations in multimodal signals. We show in \textbf{a)} the action of tapping a wooden spatula and then transitioning to a stiring motion. In \textbf{b)}, we show a state change between a raw egg and a hard-boiled egg through discrete differences in the dynamics when an impulse motion is applied while the egg is in a hand grasp. Furthermore, in \textbf{c)} We deploy a complex scenario of scene understanding through touch digitization in which a kitchen appliance and utensil are used with four artificial fingers to capture how modalities change over the course of making scrambled eggs.}
    \label{fig:fig_N1}
\end{figure*}

\subsection{Illumination System}
We describe two metric parameters for the illumination performance within the volume: background uniformity, which measures how evenly the light is distributed, and image-to-background uniformity contrast, which measures how well impressions on the fingertip's surface stand out compared to the background. 
A common approach is the embodiment of an internal structure, which serves as a hemispherical light pipe and provides fingertip rigidity. 
However, an internal light pipe structure produces illumination artifacts in the form of glint and hotspots from the convex geometry, which contribute to a degradation in image metrics. 
Prior approaches use a textured surface to reduce these artifacts using Lambertian scattering of incident light rays. 
We model the reflective layer surface properties with controlled degrees of scattering from polished to Lambertian, where the entire hemispherical surface acts as an integrating sphere, to show that a Lambertian scattering surface is not the optimal approach to achieve high performance. 
This motivates us to use a rigid solid volume instead of the more common hollow volume or the use of an internal support structure in conjunction with controlled reflective surface scattering parameters moving away from Lambertian surfaces.

\subsection{Multi-modal Sensing}

Multi-modal information capture outperforms prior visuo-tactile techniques in sensitivity to spatial and force measurements, where rapid changes in dynamics or state occur, shown Figure~\ref{fig:fig_N1}B, \ref{fig:fig_N2}B, \ref{fig:fig_2}A,C-E. While visual information provides insight into environmental and object contact, such as textures and surface deformations, this only provides a subset of fingertip-to-object-environment understanding. We further evolve the platform's capabilities to include sensitivity to non-vision-based modalities. For instance, when in contact with the environment, dynamic forces and signals are experienced, such as swiping the fingertip across a surface or the very moment a contact transient or slip occurs. We capture this information through in-fingertip audio microphones and pressure MEMS-based sensors and show the ability to determine the level of liquid inside an opaque bottle (see Figure~\ref{fig:fig_2}A) and understand the nuances of surface texture between different objects at a much higher frequency (upwards to 10 kHz) than visual capture (240Hz) (see Figure~\ref{fig:fig_2}B). We further include modalities to understand object state, which is not necessarily a function of contact but can provide priors in understanding touch through heat and smell. Such priors can estimate if an object may be slippery due to the presence of water, soap, or butter and if an action or object may present a danger to human contact due to its temperature.

\subsection{On-device AI Processing}
Inspired by the human reflex arc, where quick reactions to input stimuli on the fingertip benefit from the central nervous system instead of a round trip to the brain, we design a similar local processing response on the artificial fingertip. 
Specifically, we include within the form factor of the fingertip a neural network accelerator to process the sensory reading and allow for direct control, providing actions to a robotic end effector for controlling the phalanges of a robot finger.
While this is a new era of on-device fingertip processing, we study two main effects contributing to faster response: latency and jitter. 
Latency results from the average time required to process a signal of interest, and jitter is the variation in mean time based on system overhead, which may occur due to host processing or bandwidth constraints. 
Compared to typical methods of using an artificial fingertip with an external host, we achieve a 2x reduction in latency and jitter towards performing an action through local on-device processing.

\begin{figure*}[!t]
    \centering
    \includegraphics[width=\linewidth]{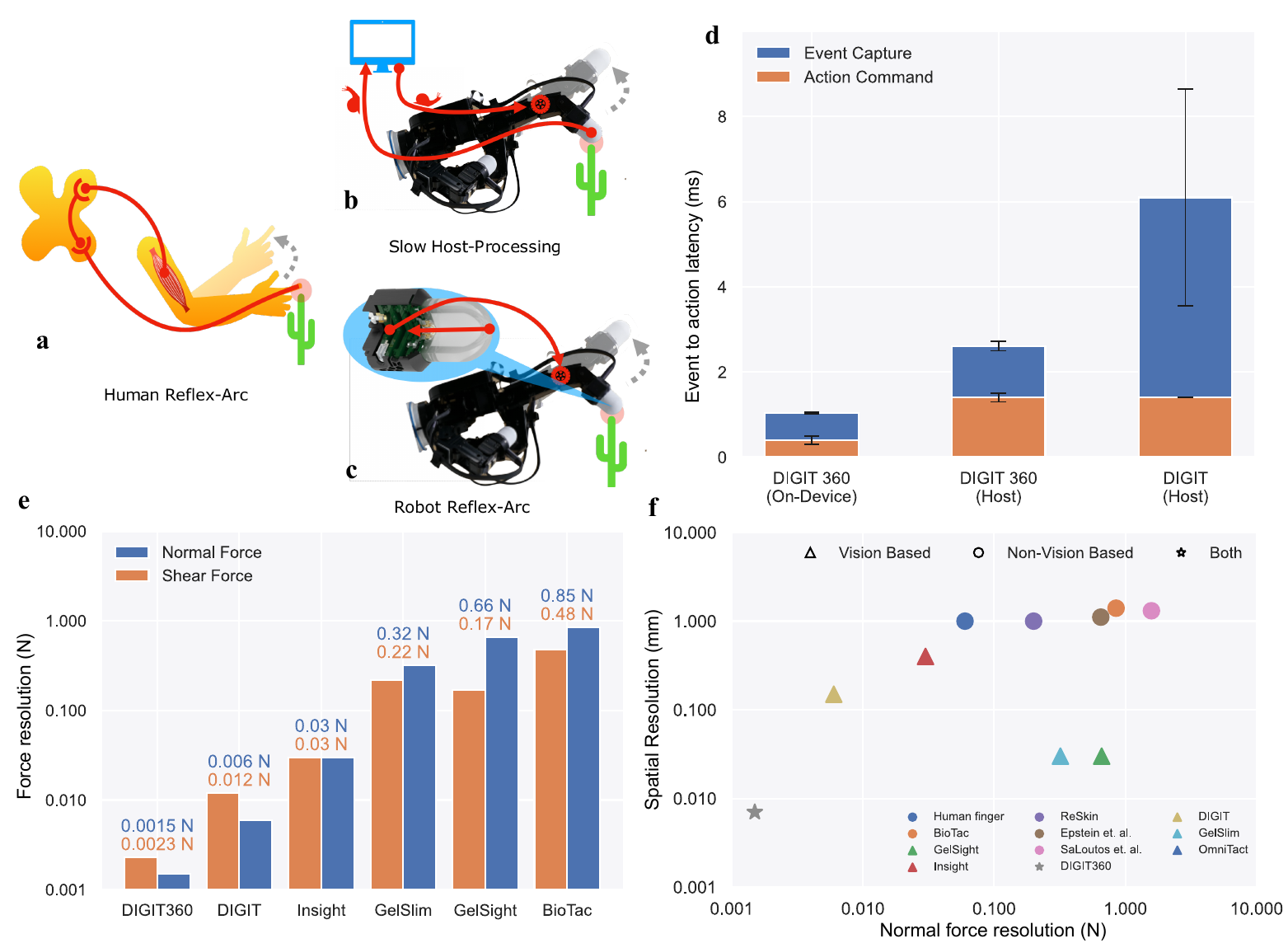}
    \caption{\textbf{a)} We introduce an analog of the human reflex arc by quickly processing sensory input within the fingertip, directly controlling the actuators of a robot hand to retract in response to touching an object.  \textbf{b)} Typical tactile processing and control paradigm transfer of sensory data to a remote computer for processing. This requires sufficient bandwidth and introduces communication latency. \textbf{c)} Local processing for mimicking the reflex arc with the system fingertip. Our system can use the on-device AI neural network accelerator for local processing to decrease overall latency between event and action. \textbf{d)} Mean and standard deviation of the event-to-action latency. The on-device local processing and control loop takes 1.2 ms compared to the traditional paradigm, which takes 2.5 ms on a Digit 360 and over 6 ms on a Digit. \textbf{e)} A comparison between normal and shear force sensitivities across devices. \textbf{f)} We compare the performance of our artificial fingertip sensor against existing sensors, and we show that our sensor delivers significant improvements (4X for spatial, 6X for normal forces, and 9X for shear forces) on all the metrics evaluated.}
    \label{fig:fig_3}
\end{figure*}

\section{Evaluation}\label{sec3_evaluation}

We evaluate the artificial fingertip, Digit 360, in performance with respect to spatial resolution, shear and normal forces, illumination, and multi-modal ablations. More details and additional experiments can be found in the methods.

We first model the fingertip surface as a two-layer stack formed by an external diffusive material adhered to an internal reflective thin film, which is grown onto the non-rigid solid silicone body of the fingertip. We then explore the effects of the non-rigid solid silicone surface mechanical properties, texture, and the degree of controlled light scattering to find an optimal performance metric between background uniformity and image contrast. We show that increasing controlled surface texture scatter from 1-degree scatter to Lambertian scatter results in an increase in background illumination uniformity, thereby increasing image impression contrast. However, with low degrees of scattering, intense hotspot artifacts dominate the background, whereas when the degree of scattering approaches Lambertian scattering, these artifacts decrease along with a decrease in image contrast, which directly results in reduced sensitivity to impression stimuli. With little or no scattering with a polished surface, it is evident that minimal background illumination is present, which motivates the production of shadows created by indentations against the fingertip surface. Furthermore, glint reflections off of produced indentations are minimal to non-existent and do not produce a consistent appearance across the surface. On the contrary, with the traditional method of visual-tactile sensors using Lambertian scattering surfaces, we show that the hemispherical sensing surface acts as an integrating sphere, where shadows cast by direct illumination striking the indentations are wiped out by scattered illumination from other areas and even while imaging may occur on far off-axis angles, their contrast is low. We introduce a controlled degree of scattering in which an optimized uniform background illumination is achieved that lends itself well to contrast between indentations and the surrounding surface; furthermore, all indentations are imaged (see Figure \ref{fig:fig_1}). 

We evaluate the normal force sensitivity and collect tuples of normal forces applied by a micro-indenter and corresponding outputs from the sensors, and then train a deep learning model from this dataset. The trained model (see Figure \ref{fig:fig_1}A) can predict the normal forces applied with a median error of 1.01 mN (Region 1). Similarly, to measure the shear force sensitivity, we collect tuples of shear forces applied by a micro-indenter and corresponding outputs from the sensors and then train a deep learning model from this dataset. The model (see Figure~\ref{fig:fig_1}A) is capable of predicting shear forces applied with a median error of 1.27 mN (Region 1). In contrast to previous sensors that required the presence of explicit markers, this result demonstrates that with a sufficiently high optical resolution, it is possible to directly use the internal texture of the elastomer to measure shear forces.

To carry out spatial resolution evaluations, we define the spatial resolution of an artificial fingertip sensor as the minimum feature size that can be resolved with a modulation transfer function (MTF) $\ge$ 0.5; this is determined by how well the contrast is preserved and quantified by line pairs per millimeter. We first simulate the imaging system from the design, which yields that on-axis contacts are resolvable for features of size $\ge$6 um for region 1, $\ge$8 um for region two resolves, and $\ge$22 um for region 3. We then validate these results by collecting data with a two-pronged micro-indenter depressed onto the fingertip, varying the distance between the two prongs, and observing the taxel intensity line profile; both the visual validation and the inspection of the taxel profile intensity confirmed it is possible to clearly distinguish features as small as $\ge$7 um for region 1 (see Figure~\ref{fig:fig_1}C).

\begin{figure*}[!t]   
    \centering
    \makebox[\textwidth][c]{
        \centering
        \includegraphics[width=\linewidth]{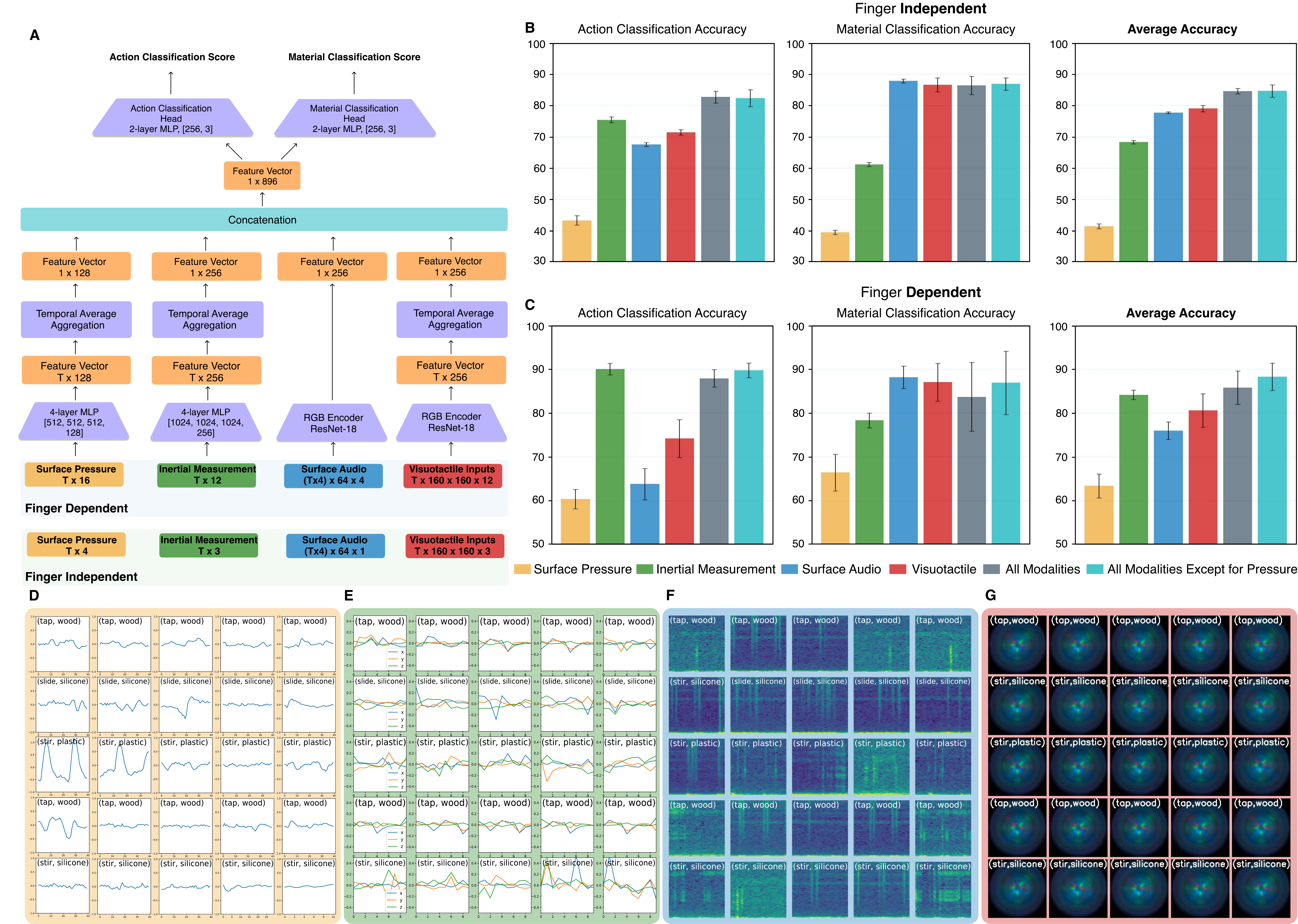}
    }
    \caption{We perform two ablations using fingertip multi-modal information, finger independent, and finger dependent neural network models where the number of modalities and contributing fingers are scaled. 
    The first, \textbf{a)} finger independent model utilizes a 1.3s sample window across all the fingers as the input to the model (T=10, visuotactile Tx120x120x3; inertial measurement 10x3; surface pressure Tx4; and surface audio (Tx4)x64x1 for each model ablation we increase the number of modalities in the dataset, 
    whereas \textbf{b)} Finger dependent, concatenates each 1.3s sample window as the input to the model (T=10 visuotactile Tx120x120x12; inertial measurement Tx12; surface pressure 10x12; and surface audio (Tx4)x64x4. Additionally, we show an implicit object encoding through the surface pressure when all four fingers are concatenated, with higher performance than the finger-independent scenario. We show a random sampling of actions and materials for \textbf{d)} surface pressure inputs, \textbf{e)} inertial measurement inputs, \textbf{f)} surface audio inputs, and \textbf{g)} visuotactile inputs.}
    \label{fig:fig_N2}
\end{figure*}

Multi-modal information such as vibrations upwards of 10 kHz, auditory clues, sensitivity to heat, and smells play an essential role in human touch \citep{Johansson1979Tactile}. However, typical vision-based tactile sensors do not contain a broad range of multi-modal capabilities to capture this information or operate lower sensing frequencies, such as 60 Hz. Even with Digit 360’s fast camera, which operates at 240 Hz, highly dynamic movements may not be fully captured.

Multi-modal touch digitization can be characterized by two studies: a) a single modality based characterization and, b) a holistic characterization of all the modalities collectively.

\textbf{Single modality based characterization:} We first look at an approach where each modality is characterized by its own respective performance. We evaluate capturing vibrations up to 10 kHz, which can distinguish between different materials with light swiping motions of the finger. Furthermore, we show that these multi-modal features can be used to detect the amount of liquid inside a bottle by simply tapping it with a fingertip (see Figure \ref{fig:fig_2}A), like audio and vibratory clues consumed by humans during object interactions.
Along with audio and vibratory clues, humans evaluate touch interactions based on changes in local heat gradients. We show that we can detect changes in heat gradients that reflect object state: room temperature, warm, hot, dangerous (see Figure~\ref{fig:fig_2}C). In regards to object state, a limited amount of information is captured by the vision-based modality during contact. We employ the use of local gas sensing at each fingertip to understand nuances in object state, for example, determining if the object is slippery or wet. We show that with this modality, we can sense these parameters during approach and contact (see Figure~\ref{fig:fig_2}E). Specifically, we evaluate contact different samples, which not only provide gas signatures but local environmental information such as humidity and temperature gradients to distinguish between two similarly looking liquids and coffee or coffee grinds (see Figure \ref{fig:fig_2}D). Multi-modal sensing complements the primary vision-based sensing modality and enables future research into the importance of the different touch modes for task-specific applications.

\textbf{Multi-modality based characterization:} Digitizing the multimodal signals from a sequence of touch events while deploying everyday actions such as tapping objects against one another or sliding an object within the surrounding environment shows distinctive features across several modalities, as shown in (Figure \ref{fig:fig_N2}A). We set up a data collection scenario with a robotic arm and a 4-finger robotic hand to continuously act as an action platform. Observed are different signal modalities between tapping and stirring motions of a wooden spatula against a kitchen pan; however, in the case of a more complex and compound task such as making scrambled eggs, only some modalities provide strong discernable cues as to the dynamics of the scene, (Figure \ref{fig:fig_N2}C, acceleration, surface pressure, temperature). In contrast, other modalities suggest the state of the objects in the scene ("Are the eggs done cooking?") shown in (Figure \ref{fig:fig_N2}C, gas signature, humidity). While this is a visually-aided differentiation between actions, we deploy a deep neural network (DNN) model to determine the cross-modal significance of each modality in classifying actions and characteristic intrinsics, such as the object material being used in that action. The DNN is based on captureing three categories of touch signals, those from low-frequency temporal, such as surface pressure and intertial measurement, high-frequency temporal such as surface audio and finally visuotactile inputs.

On the design of the DNN as shown in (Figure \ref{fig:fig_N2}A), first, we design a modality-specific encoder for each modality. Second, we process the temporal modalities with a multilayer perceptron (MLP) network. In contrast, the vision and surface audio modalities are processed with a ResNet-18 network without pre-training, acting as an RGB image encoder. The outputs of the MLPs and RGB encoders are used as feature encoders, which are concatenated to a final feature vector, and two final output MLP heads provide action and material classifications. We record a dataset composed of four active fingers digitizing a specific action-material pair task. This dataset contains a total of $\approx$ 620k samples across all four fingers. 

Two paradigms are presented and trained: a finger-independent (Figure~\ref{fig:fig_N2}B)  where each finger is treated as a separate sample, and a finger-dependent model (Figure~\ref{fig:fig_N2}~C) in which all fingers are concatenated into a single sample for input into the model. These two ablations are tested to show the cross-modal dependence; in the case of finger-independent, the intertidal and pressure-based modalities perform worse than in the case of finger-dependent where there is an implicit encoding of the multi-finger and hand grasp on the object.
We evaluate the finger-dependent model on the assumption that the contributions of each modality define a pose-dependent cross-modal contribution across the fingers. 

Table~\ref{tab:multimodal-action-material-cls}) shows baseline performance. Each modality is sampled with a window size of 1.33~seconds as shown in (Figure~\ref{fig:fig_N2}D-G). With this data-driven approach, we show that inertial measurements predominantly dominate action classification, while material classification is dominated by visuotactile and surface audio signals. Furthermore, comparing finger-dependent and independent scenarios, cross-modal effects are established based on the collective contributions of all the fingers, where all modalities together provide higher classification accuracies than the finger-independent scenario. Additionally, we show that by increasing the number of modalities used in both action and material classification, the performance of the classification network increases. 

Finally, inspired by the human reflex arc \citep{Dewey1896reflex}, we demonstrate a fast reflex-like control loop using the on-device AI neural network accelerator for local processing. Compared to a Digit sensor using an external computer for processing, on-device processing on Digit 360 reduces latency from 6 ms to 1.2 ms (see Figure~\ref{fig:fig_3}D). With the increasing computational power of on-device accelerators, larger touch sensing surface areas, and the increasing use of AI models for touch processing \citep{lambeta2021pytouch} – the capability of processing data locally and transferring only high-level features will prove crucial for touch processing.

\section{Discussion}\label{sec4_discussion}

We show design principles that advance the state of artificial fingertip sensing toward digitizing fingertip interactions between the environment and objects. We design an artificial fingertip that is more sensitive in spatial and force sensitivity compared to similar methods with the additional benefit of multi-modal sensing features and a local processing ability. Our results demonstrate the digitization of touch with capabilities that outperform a human fingertip. We believe the richness of touch digitized by our modular research platform Digit 360 opens new promising venues for studying the nature of touch in humans and investigating key questions around the digitization and processing of touch as a sensor modality \citep{Hayward2011Is}. Moreover, this technology opens the doors to broader adoption of touch sensors beyond traditional niche research fields: In robotics, to improve sensing and manipulation capabilities with benefits for applications in manufacturing and logistics, medical robotics, agricultural robotics, and consumer-level robotics; In artificial intelligence, to investigate the learning of appropriate tactile and multi-modal representations, and corresponding computational models that can better exploit the active, spatial and temporal nature of touch. Further potential applications include virtual reality, telepresence, teleoperation, prosthetics, and e-commerce. To support and foster research in the exciting fields of touch sensing and processing, we open-source Digit 360’s design at \url{https://github.com/facebookresearch/digit360}.

\clearpage

\section{Methods}\label{sec11}

\renewcommand{\subsection}[1]{\textbf{#1}}
\renewcommand{\paragraph}[1]{\textit{#1}}
\def\percent{\%}

\begin{figure}[!t]
    \centering
    \begin{minipage}[b]{0.45\textwidth}
        \centering
        \includegraphics[width=0.45\linewidth]{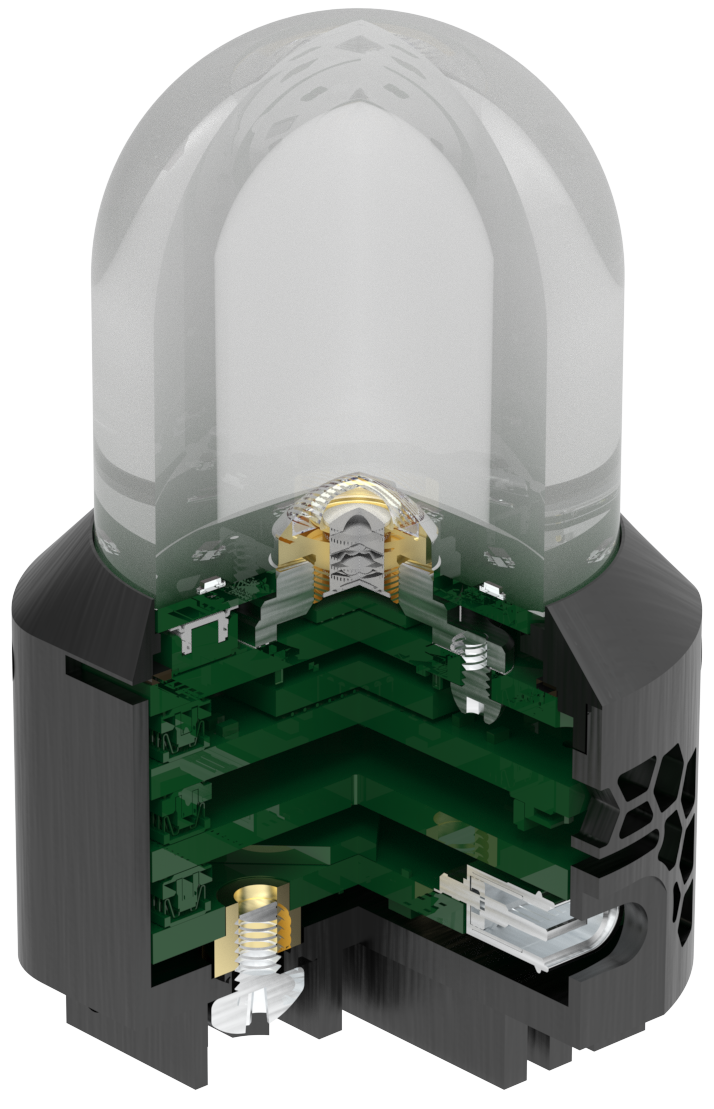}
        \includegraphics[width=0.45\linewidth]{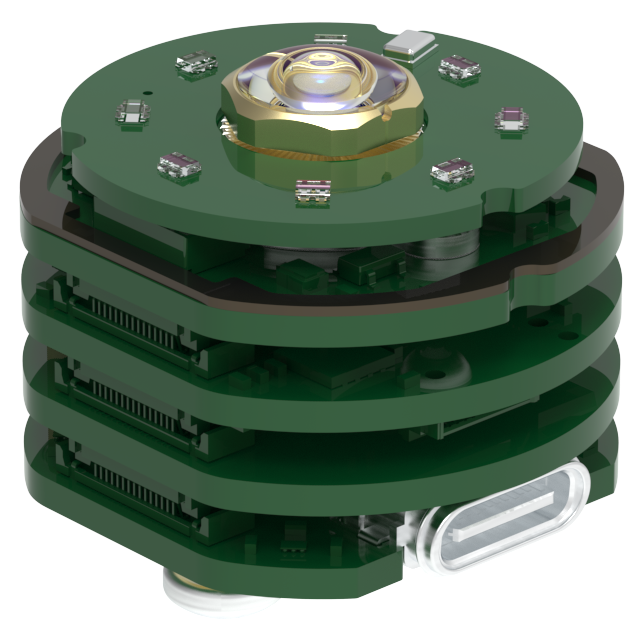}
        \caption{Cutaway diagram of the \sensor{} and its modular electronics.}
        \vspace{2em}
        \label{fig_d360_design}
    \end{minipage}
    \hspace{0.05\textwidth} 
    \begin{minipage}[b]{0.45\textwidth}
        \centering
        \begin{minipage}[t]{0.5\linewidth}
        \centering
        \subcaptionbox{Fingertip\\Regions}
          {\includegraphics[origin=c,width=0.75\linewidth]{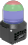}}%
      \end{minipage}%
      \begin{minipage}[b]{.5\linewidth}
        \centering
        \subcaptionbox{Lens}
          {\includegraphics[origin=c,width=0.8\linewidth]{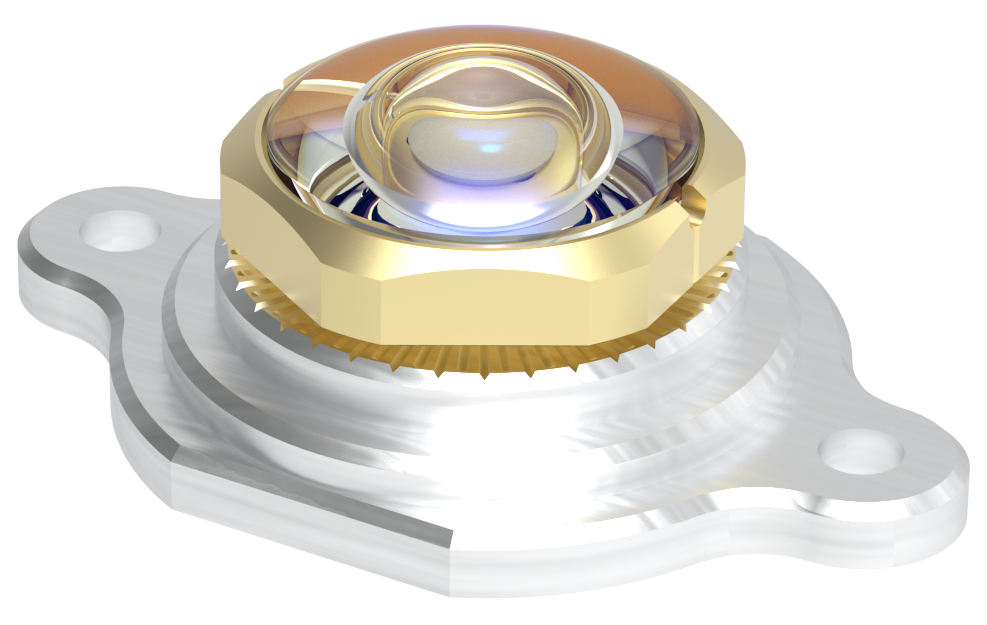}}
        \centering
        \subcaptionbox{Lens\\Cutaway}
          {\includegraphics[origin=c,width=0.8\linewidth]{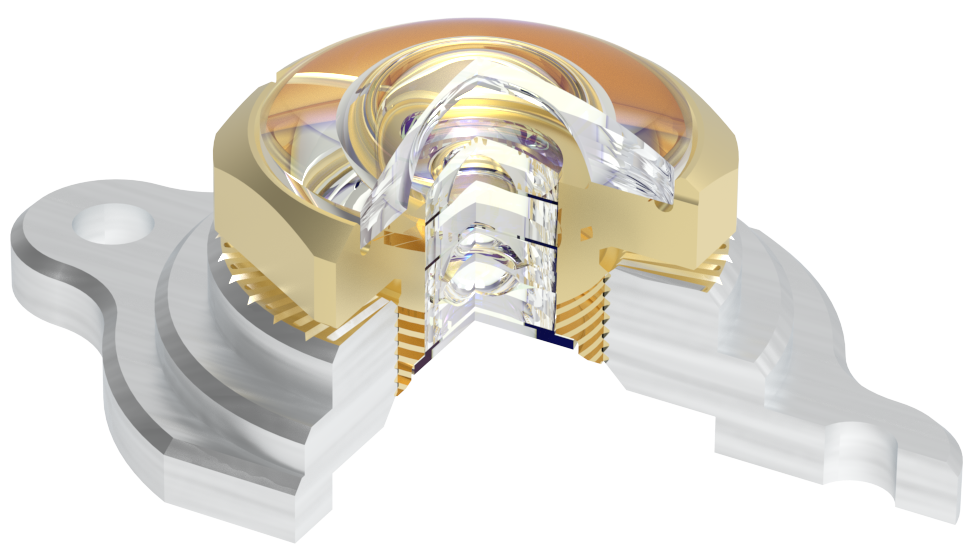}}%
      \end{minipage}%
      \caption
        {%
          Hyperfisheye lens designed specifically for capturing omnidirectional tactile sensing images. Three regions of interest are selected for optical system performance, tip (red), prominent contact surface (green), and base (blue).
          \label{fig:sensor-regions-lens}%
        }%
    \end{minipage}
\end{figure}

\subsection{Platform architecture}
We developed this platform specifically for advancing the state of the art in tactile sensing research. As well, it is a platform based on modular principles to provide an omnidirectional vision-based, multi-modal sensing system, and on-device AI capabilities. We achieved a modular design by isolating each part of the system into a small surface area electronic assembly, as shown in Figure~\ref{fig:fig_1}. By facilitating the reuse of each sub-system, we present a novel solution to advance tactile sensing research by reducing the complexity and the design and manufacturing iteration cycles. This enables researchers to modify the overall system by adding, removing, and changing sub-systems rather than having to design a stack of new hardware to support changes to system design. Furthermore, this modular system architecture allows for selecting combinations of sub-systems to facilitate the introduction of newer technologies. Feature removal reduces costs when adapting the system to new mechanical form factors. Also, the possibility of replacing the sensing fingertip allows adaptability for different environments and tasks - for example, by using different sensitivities and stiffness in the fingertip material or by using fingertips with markers to introduce more prominent optical-flow features. Based on simulation to achieve optimal spatial and force resolution, we were able to introduce a novel process for the design and manufacturing of the sensing fingertip.

\begin{figure}[!t]
    \centering
    \begin{minipage}[b]{0.38\textwidth}
        \centering
        \includegraphics[width=\linewidth]{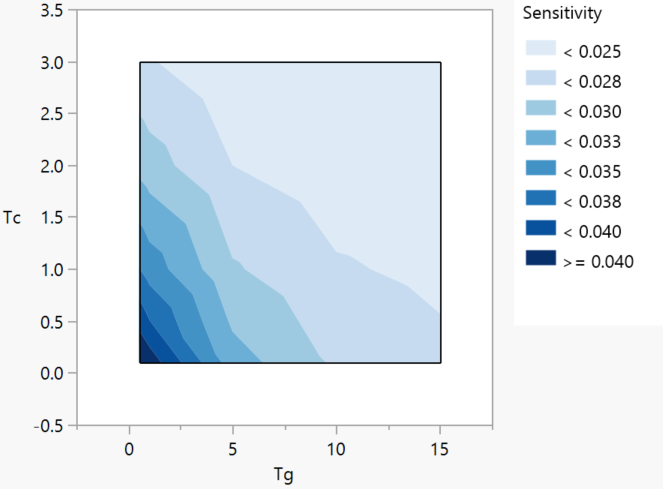} 
        \caption{Touch sensitivity surface map plot of $E_c = 5.0$ and $E_g = 3.0$ \SI{}{\mega\pascal}. We show that sensitivity increases when gel and coating thickness are decreased.}
        \label{fig:fem-coating-thickness}
    \end{minipage}
    \hspace{0.02\textwidth} 
    \begin{minipage}[b]{0.55\textwidth}
        \centering
        \includegraphics[width=\linewidth]{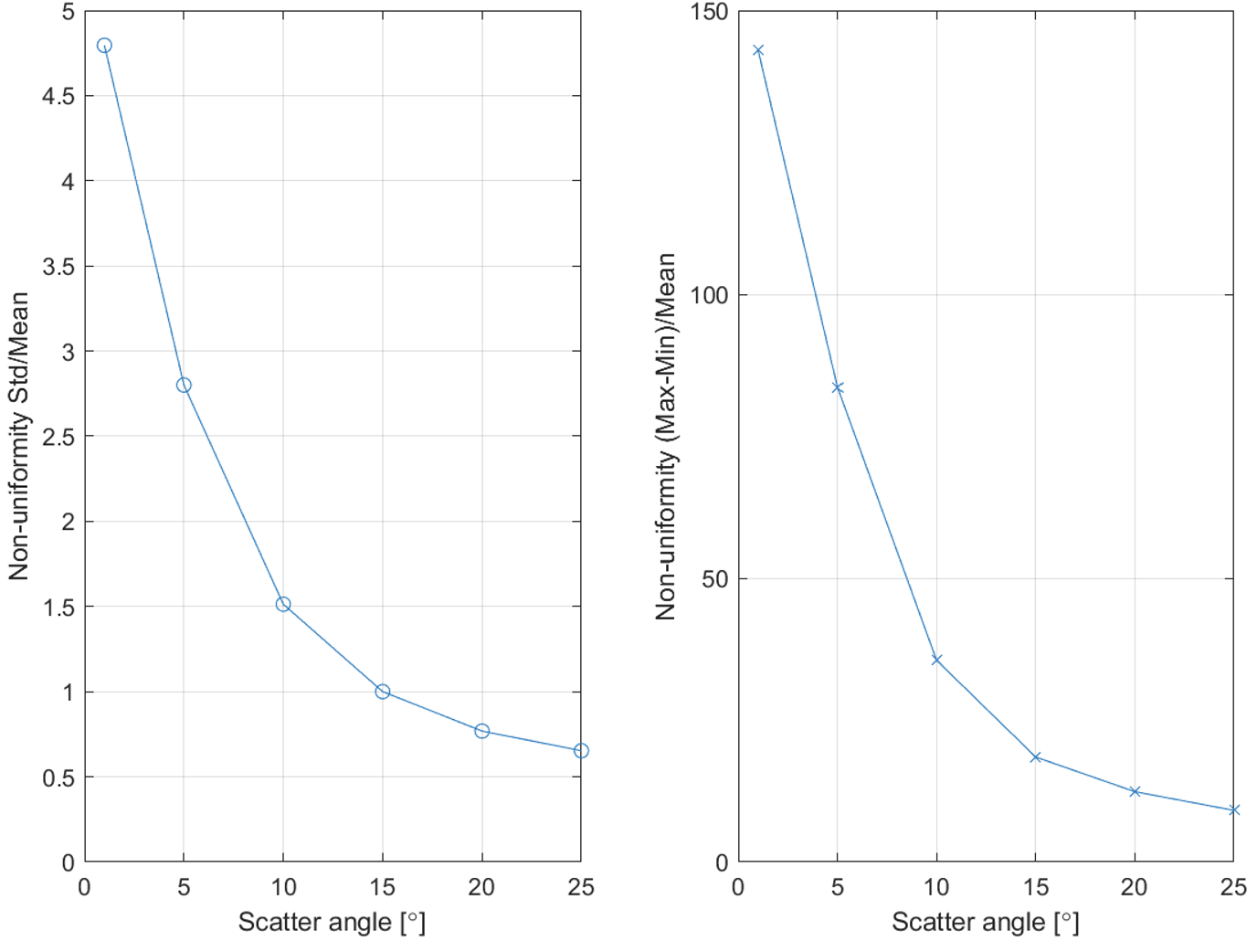}
        \caption{Minimizing the background non-uniformity with increasing scatter angle, thereby improving the detection of surface indentations.}
        \label{fig:scattering-metrics}
    \end{minipage}
\end{figure}

\begin{figure}[t]   
    \centering
    \includegraphics[width=\linewidth]{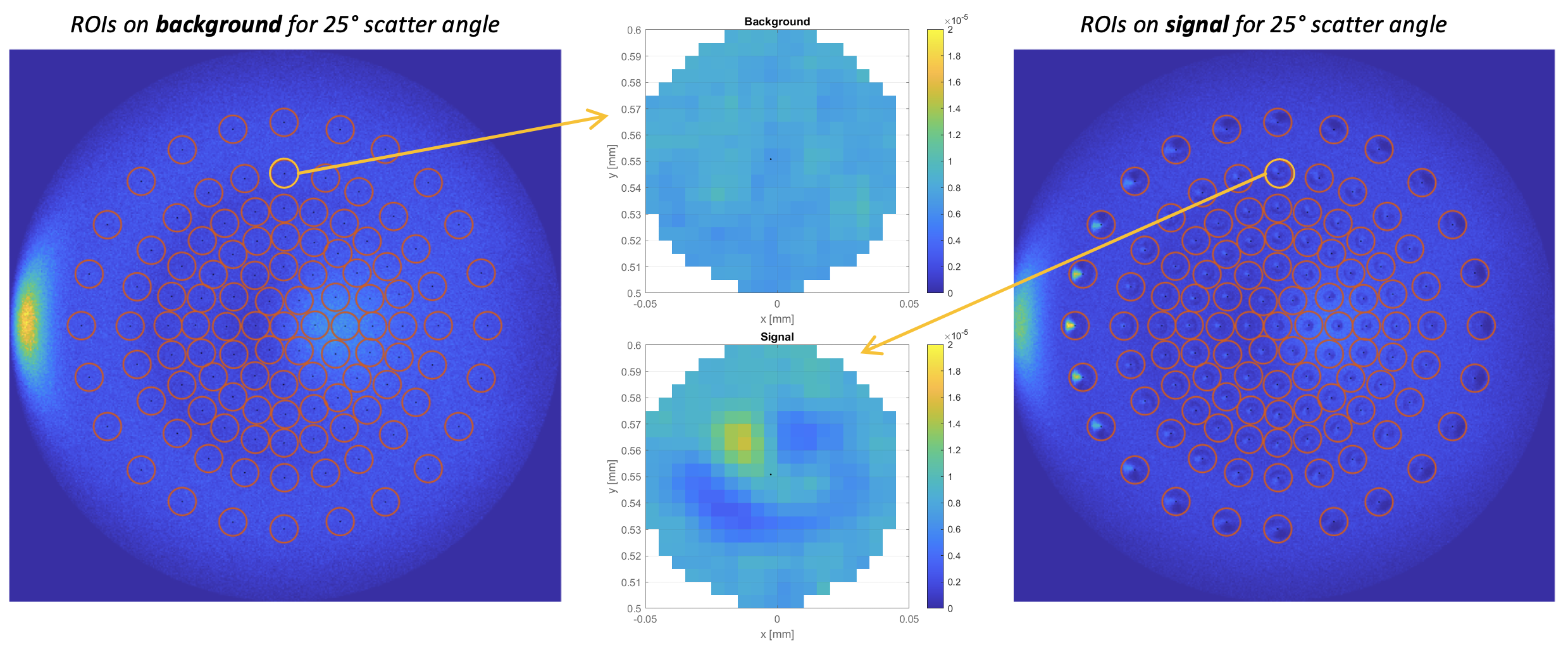}
    \caption{Selecting regions of interest for background and indentation with multi-contact indentations across the surface of the fingertip.}
    \label{fig:roi-contrast-to-noise}
\end{figure}

\begin{figure}[t]   
    \centering
    \includegraphics[width=\linewidth]{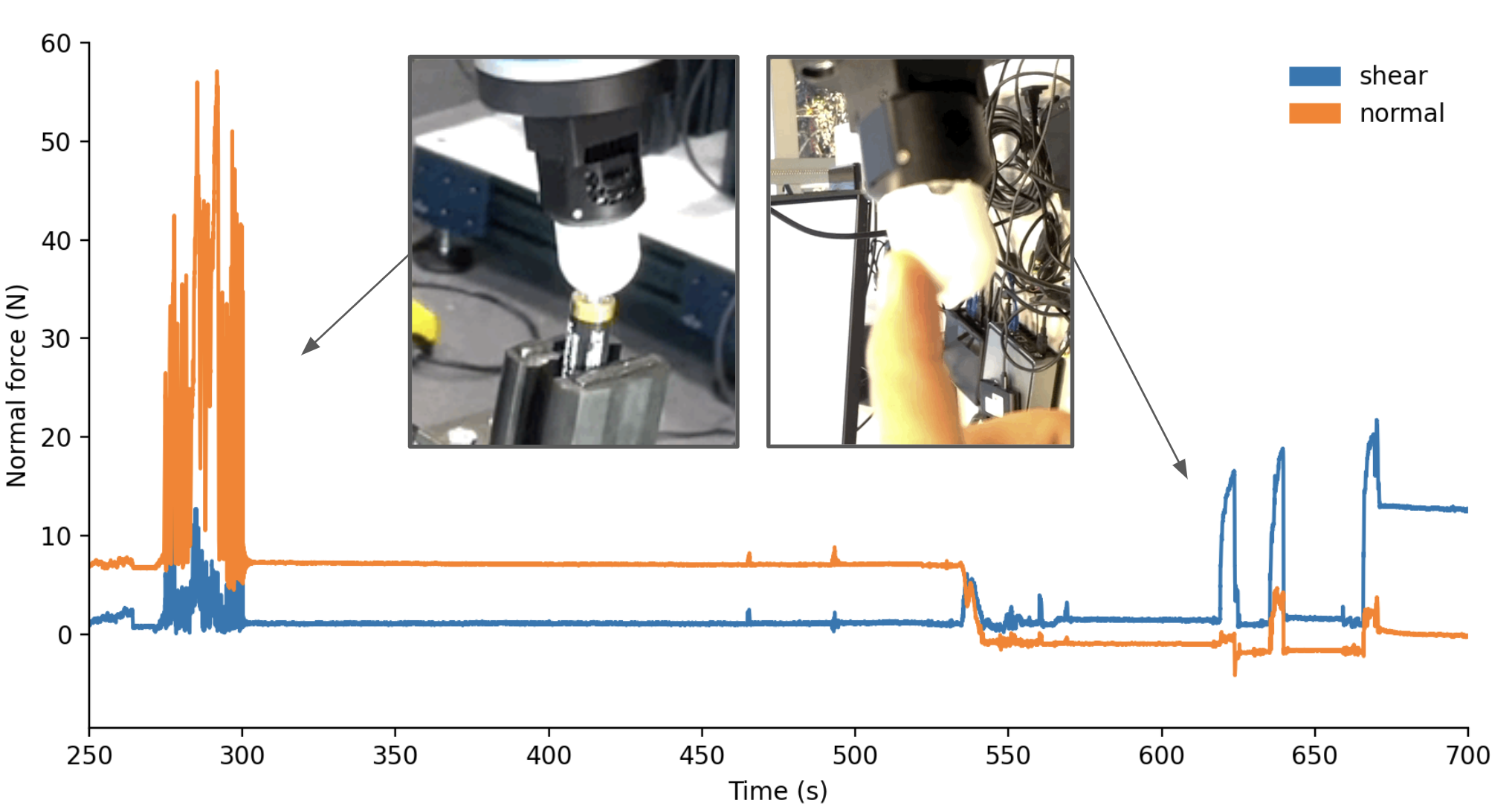}
    \caption{The maximum normal and shear force applied to the artificial fingertip surface prior to detachment from the main body. We apply an incremental normal and shear force to the body of the soft artificial fingertip and record the ground truth force data from a stationary force-torque sensor. When an abrupt change occurs, this indicates that maximum force has been reached.}
    \label{fig:max-force}
\end{figure}

\begin{figure}[t]   
    \centering
    \includegraphics[width=0.4\linewidth]{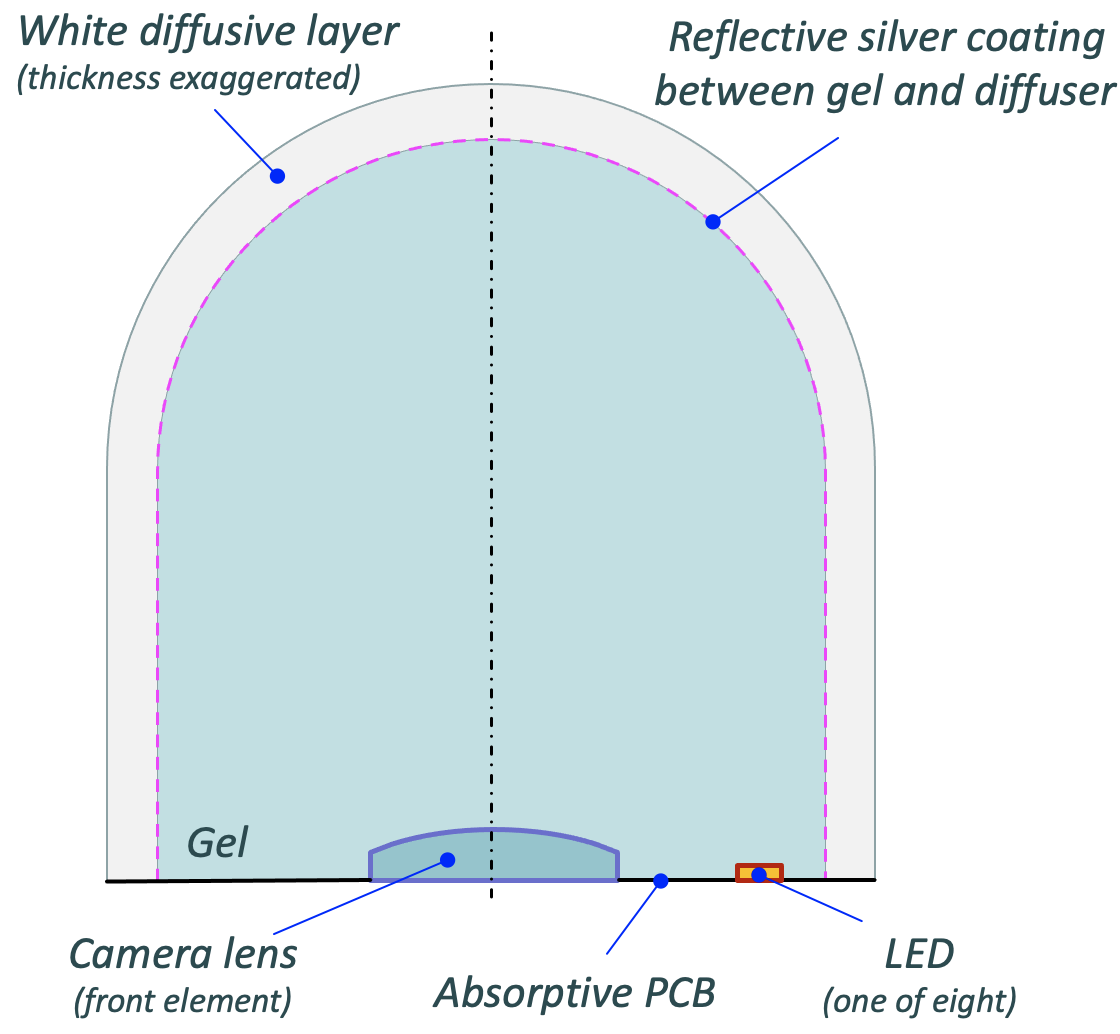}
    \hfill
    \centering
    \includegraphics[width=0.55\linewidth]{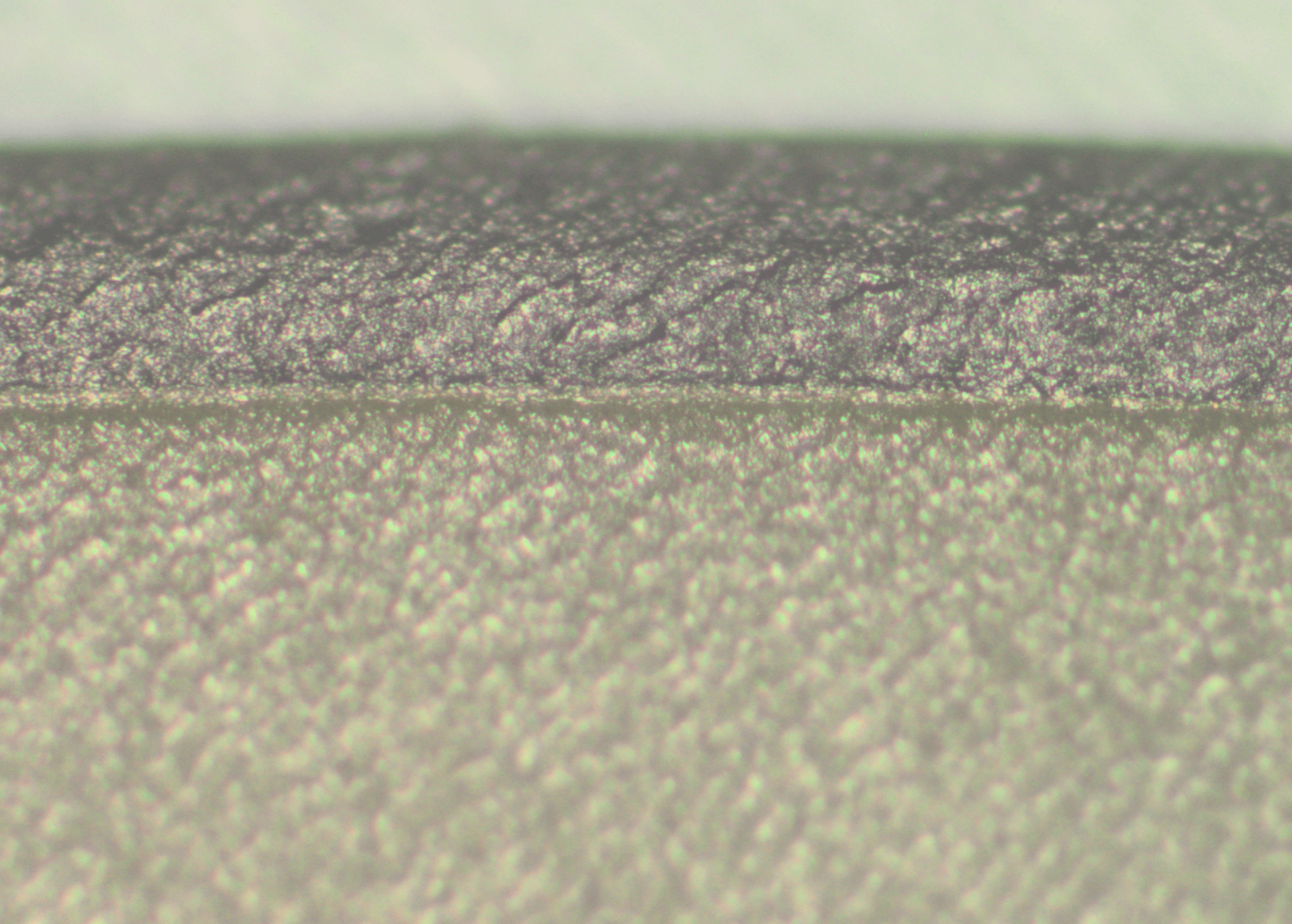}
    \caption{Cross-section of the fingertip gel coating showing the three layers: outer, silver, and base gel}
    \label{fig:gel-layers}
\end{figure}

\begin{figure}[t]   
    \centering
    \includegraphics[width=0.5\textwidth]{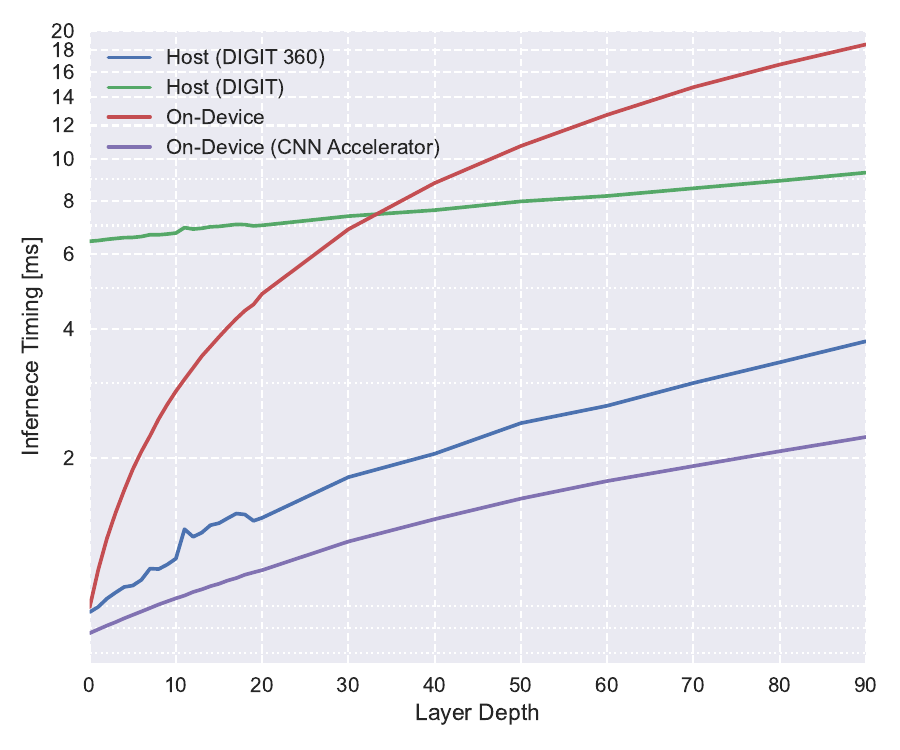} 
    \caption{Inference latency measurements comparing on-device to host incorporating the tactile data pre-processing and transfer stages. We show that using the on-device CNN accelerator improves inference time and reduces the total latency to producing actions for an MLP deep neural network.}
    \label{fig:edgeai-data-timing-mlp}
\end{figure}

\begin{figure}[t]   
    \centering
    \includegraphics[width=0.9\linewidth]{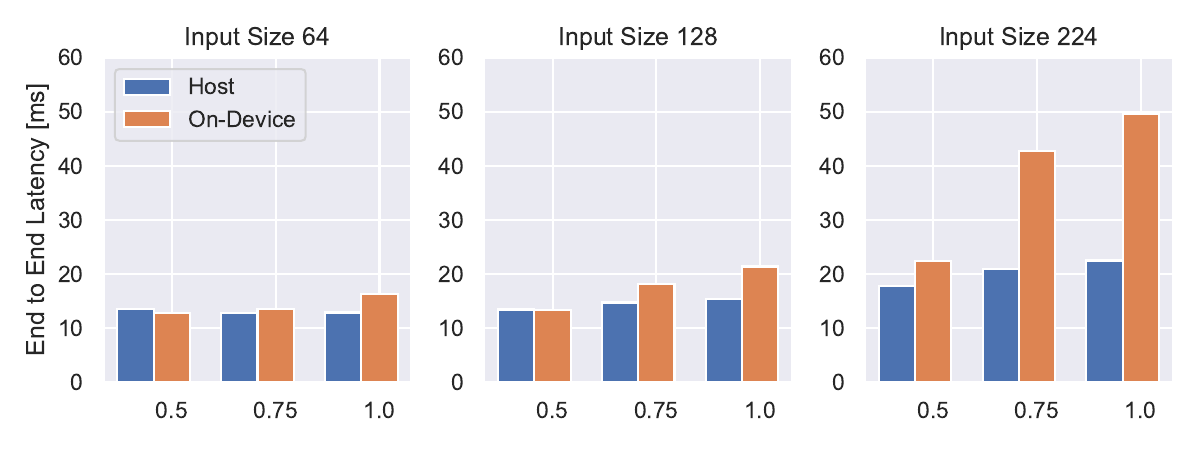} 
    \caption{We show the average pipeline latency for a MobileNetV2 network from acquiring tactile data, transferring data, pre-processing, and inference to providing actions for varying image and channel width sizes}
    \label{fig:edgeai-data-timing-mobilenet}
\end{figure}

\begin{figure}[t]   
    \centering
    \includegraphics[width=\linewidth]{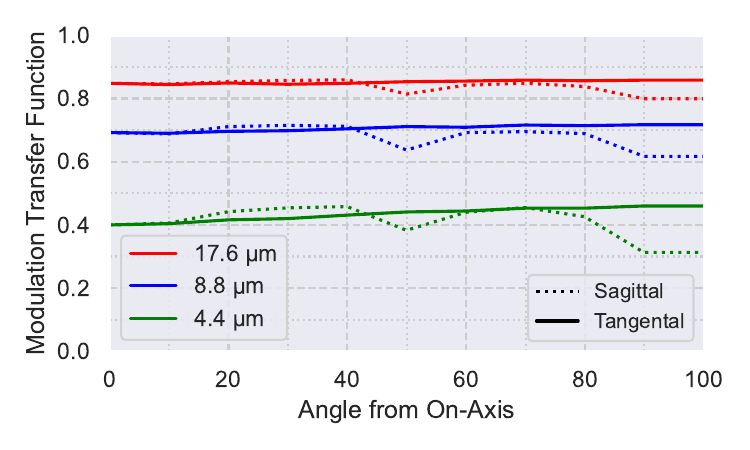} 
    \caption{We show the simulated performance of the \sensor{} vision system from on-axis to far-field contact for increasing line pairs per millimeters translated to spatial resolution for sagittal and tangential responses.}
    \label{fig:spatial-mtf-ideal}
\end{figure}

\subsection{Fingertip sensitivity simulation to input stimuli}
A 3D Finite Element Method (FEM) model using Comsol Multiphysics was developed for analyzing and characterizing the fingertip material stack-up. The goal was to identify the sensitivity and resolution of the sensor. First, a FEM model was developed to identify the key parameters that presented the most significant change in sensitivity and resolution. Since the fingertip is isotropic and revolves around the origin, only a quarter of the sensor was modeled for faster computation using a multi-layer-based model. The multi-layer model comprises the base gel, polymer, and coating layers.

We used a Hysitron TI 980 TriboIndenter for nano-mechanical characterization of the fingertip polymer Young’s modulus, E. This system has in-situ high-resolution imaging, dynamic nanoindentation, and a high-precision motion stage with high-resolution force-sensing tips. For characterization, a 30 µN force was applied with a 10 um probe tip. The corresponding force-displacement curve was measured, yielding Young’s modulus of E = 2.86 MPa. Using the experimental E value, the FEM models were updated to correct the simulations. Furthermore, the same force value applied and maximum displacement, Dmax, was measured, resulting in the verification of simulation, Dmax = 2.1 um and experimental Dmax = 2.2 um measurement results, with error $\le$ 5\percent. Additionally, multiple measurements were taken across varying samples of the fingertip. An average value for E was measured at E = 2.6$\pm$0.74 MPa. Both Emean and Estd were used in detailed analysis for the total E range in the FEM models.

\begin{table}[t]
\centering
\setlength{\tabcolsep}{3pt}
\renewcommand{\arraystretch}{1.25}
\begin{tabular}{l|ll}
 & Host {[\SI{}{\micro\second}]} & On-Device {[\SI{}{\micro\second}]} \\ \hline
Data acquisition &  \multicolumn{2}{c}{Modality/Frequency Dependent} \\
Data transfer & 1600 & 248 \\
Sub-sampling & 6 & 393 \\
Inference & \multicolumn{2}{c}{Model Dependent}  \\
Action transfer & 530 & 40 \\
Action & 1010 & 2 \\ \hline
Total & 3146 & 683
\end{tabular}
\caption{Average data pipeline timing for host and on-device processing. We observe a $\sim4x$ decrease in action latency for dynamic tasks that involve high-velocity movements, thereby reducing the total time to action to less than \SI{1}{\milli\second}.}
\label{tab:on-device-ai-timing}
\end{table}
We further employed design-of-experiments techniques used to identify critical parameters affecting sensor sensitivity. Six different parameters were used: Rgel (gel fingertip radius), Tc (coating layer thickness), Tg (gel layer thickness), h (height), Ec (coating Young’s modulus), and Eg (gel Young’s modulus). Entertaining a full factorial design of 6 parameters would lead to 64 models, thus a quarter-factorial design method was used to reduce the design into 16 models. Analysis of variance and prediction analysis resulted in Ec and Eg as the main effects and interactions with coating and gel thicknesses. Hence, the parameters h (height) and Rgel (gel fingertip radius) were removed from the model. To analyze the effect of gel and coating thickness, Tg, Tc on the sensor performance, design-of-experiments methods were used through sweeping Young’s modulus parameters, Ec, Eg, and thickness parameters, Tg and Tc. For the protective fingertip layer’s Young’s modulus, values of Ec,g = {0.5, 1.0, 3.0, 5.0} MPa were used. For the values of coating thickness, Tc = {0.1, 0.5, 1.0, 2.0, 3.0} mm, and for gel thickness, Tg = {0.5, 1.0, 5.0, 10, 15} mm were used. The result of the FEM analysis was visualized with surface map graphs, as shown in Figure~\ref{fig_d360_design}. Where the x-axis shows gel thickness, Tg values, and the y-axis shows coating layer thickness, Tc for each combination of coating and gel Young’s modulus, Ec and Eg. Figure~\ref{fig_d360_design} indicates that the material Young’s modulus contributes significantly to yield a minimum required Tc and Tg for high performance. A similar FEM analysis was performed for Digit. A multi-layer-based model was used. We analyzed, Tg (gel thickness), Tc coating thickness, Ec (coating Young’s modulus) and Eg (gel Young’s modulus). A parametric study was performed with Tg 0.5 to 15mm with 0.25 steps and Tg 0.1 to 3mm with coating Young’s modulus Ec 5MPa and gel’s Young’s modulus Eg 1MPa. The result of this study shows that gel and coating thickness interact, and the combination of the coating and gel thickness affects the sensor's overall sensitivity.

We determine the average and maximum forces that can be applied to the fingertip before the soft hemispherical surface is damaged. We fix the sensor to a force-torque sensor and apply an increasing force until the fingertip detaches from the body, which occurs at 40N and 20N for normal and shear forces, respectably, as seen in Figure \ref{fig:max-force}.

\begin{figure}[t]   
    \centering
    \includegraphics[width=\linewidth]{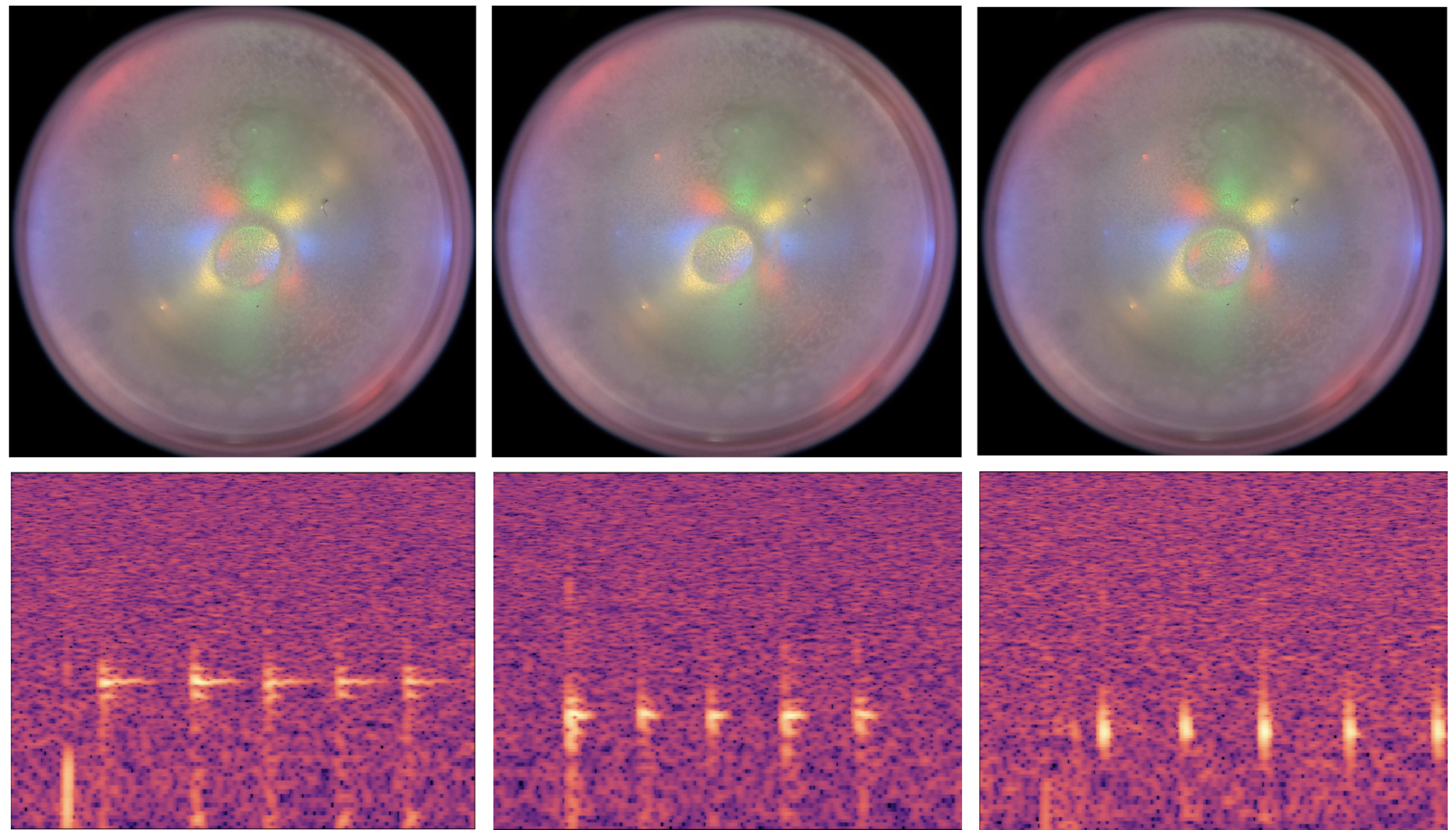} 
    \caption{Comparing touch information from \sensor{} tapping a water bottle with varying volumes of water from empty, half filled, and full. While the vision output from the sensor looks nearly identical, with only variations in the location of the touch, multi-modal data provides a larger insight into object properties beyond texture.}
    \label{fig:mm-water}
\end{figure}

\subsection{Fingertip material characterization using DMTA}
Reflecting on FEM simulations, Young’s modulus is a critical parameter in sensor performance to input stimuli and requires precise controller measurement. In addition to nanoindenter measurement, which is the point-based measurement, a set of dynamic mechanical and thermal analysis (DMTA) measurements were performed to obtain the global Young’s modulus of the gel. With this method, we measured the viscoelastic properties of polymers. During DMTA measurements, an oscillating force was applied to the material, and its response was recorded to calculate the viscosity and stiffness of the material. The oscillating stress and strain measurements are essential in determining the viscoelastic properties of the material. 

\begin{figure}[t]   
    \centering
    \includegraphics[width=\linewidth]{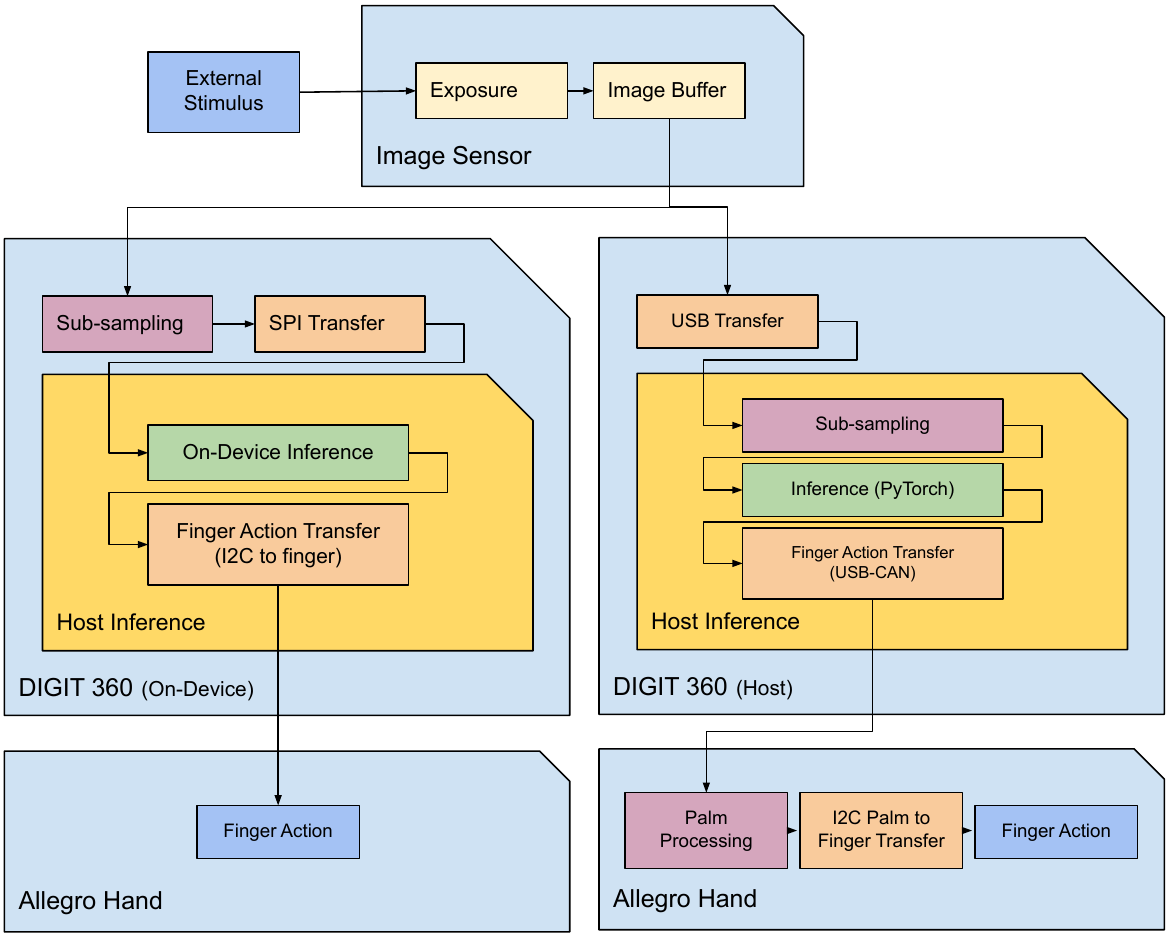} 
    \caption{Data capture pipeline for the \sensor{} vision system, touch information from external stimulus.}
    \label{fig:edgeai-data-pipeline}
\end{figure}

When oscillating force was applied, sinusoidal stress and strain values were measured. The phase difference between sinusoidal stress and strain provided information about the viscous and elastic properties of the material. Ideal elastic systems have a 0 $\degree$C phase angle, while viscous systems have a phase angle of 90 $\degree$C. Additionally, the elastic response of a material is similar to storage energy and is captured by storage modulus, while the viscous response can be considered as loss of energy, captured by loss modulus. Thus, the overall modulus of the viscoelastic material is the combination of elastic and viscous components; in other words, the summation of storage modulus and loss modulus. Another value, tan $\delta$, is used to compare the viscous and elastic modulus. DMTA measures the change in the elastic modulus, loss modulus, and tan $\delta$ with respect to temperature. As the viscosity of the material is affected by temperature and time, DMTA experiments are usually performed at different temperatures and frequencies. We use a common approach to select the operational conditions of the materials. Hence, for ideal sensitivity, the fingertip is expected to be used at room temperature and low frequency. As such, DMTA measurements were taken at 25 $\degree$C with a frequency of 5 Hz. 

\begin{table}[]
\centering
\setlength{\tabcolsep}{3pt}
\renewcommand{\arraystretch}{1.25}
\begin{tabular}{l|cccc}
\multicolumn{1}{l|}{Polymer} & \multicolumn{1}{l}{Shore} & \multicolumn{1}{l}{\begin{tabular}[c]{@{}l@{}}Storage \\ Modulus\end{tabular}} & \multicolumn{1}{l}{\begin{tabular}[c]{@{}l@{}}Loss \\ Modulus\end{tabular}} & \multicolumn{1}{l}{$\tan{\delta}$} \\ \hline
Sorta Clear 12 & 12 & 31.2 & 2.9 & 0.09 \\
Sorta Clear 18 & 18 & 65.1 & 8.2 & 0.12 \\
Solaris & 15 & 38.5 & 1.8 & 0.048 \\
Encapso K & 33 & 56.1 & 6.6 & 0.11 \\
Ecoflex Gel & 32 & 0.66 & 0.41 & 0.61 \\
Ecoflex 0010 & 10 & 1.11 & 0.32 & 0.29 \\
Ecoflex 0020 & 20 & 0.92 & 0.15 & 0.16 \\
Ecoflex 0030 & 30 & 1.17 & 0.13 & 0.11 \\
Ecoflex 0035 & 35 & 2.3 & 0.14 & 0.06 \\
Ecoflex 0050 & 50 & 1.55 & 0.19 & 0.12 \\ 
Ecoflex 0045 & 45 & 5.7 & 0.43 & 0.07 \\ \hline
Solaris A:B 0.9 :1 & \multicolumn{1}{l}{N/A} & 30.7 & 1.24 & 0.04 \\
Solaris A:B 0.8 :1 & \multicolumn{1}{l}{N/A} & 28.3 & 1.05 & 0.037
\end{tabular}
\caption{DMTA measurements performed on typical polymers used in fingertip gel manufacturing.}
\label{tab:dmta-measurements-polymer}
\end{table}

During fingertip manufacturing, different combinations of polymers with varying shore values were evaluated. To identify the global Young’s modulus and effect of varying gel mixtures, DMTA measurements were done at a sample temperature of 25 $\deg$C with a indentation frequency of 5 Hz, as shown in Table~\ref{tab:dmta-measurements-polymer}. Fingertip materials with lower Young’s modulus were preferred to optimize for higher sensitivity. We selected a fabrication using Smooth-On Solaris A:B 0.8:1 ratio for the gel fingertip base material and a Smooth-On Ecoflex 0010 thin protection layer for the thin-film layer.

\begin{figure}[!t]
    \centering
    \begin{minipage}[b]{0.3\textwidth}
        \centering
        \includegraphics[width=\linewidth]{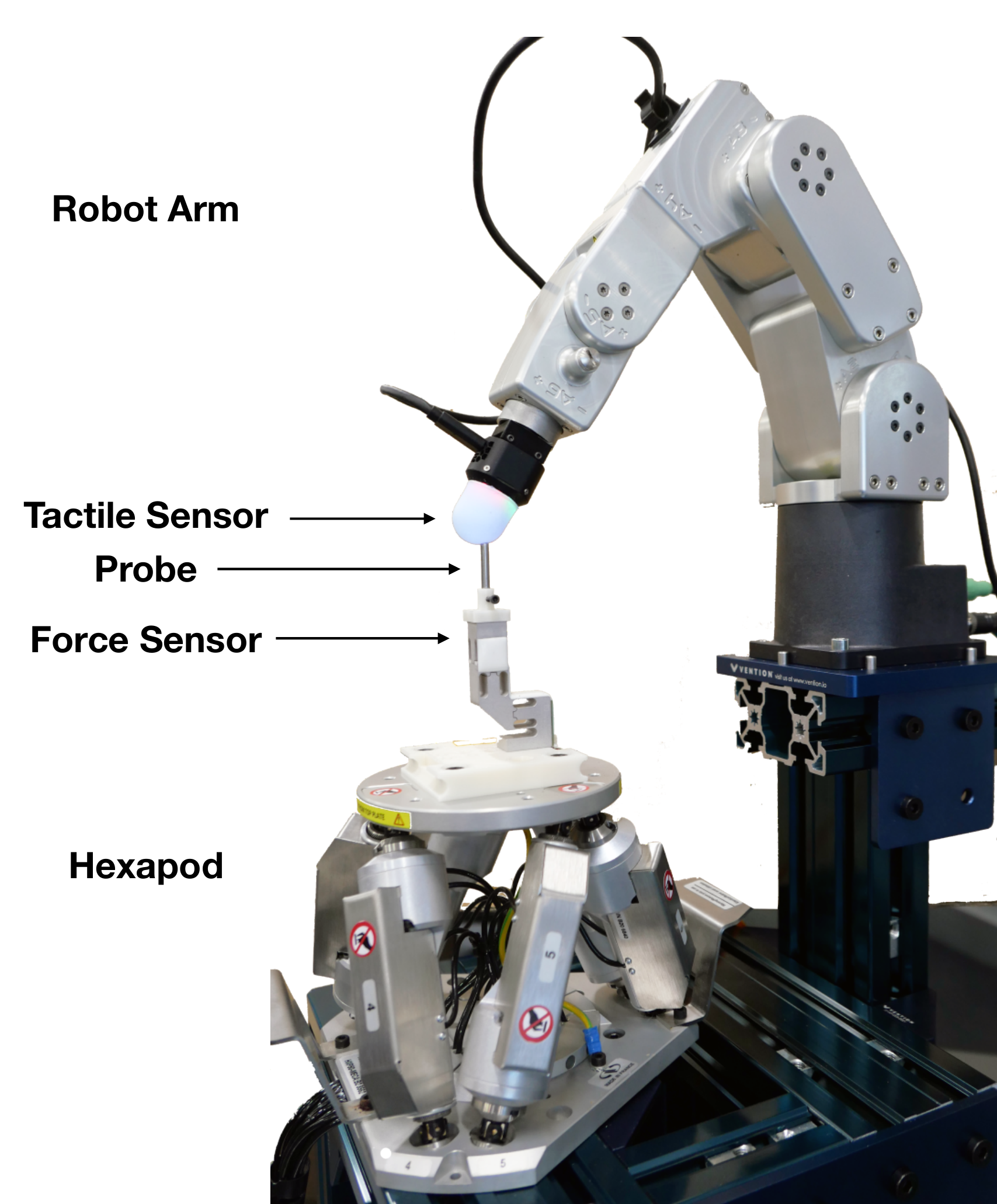} 
        \caption{We designed a 6 DoF Robot indenter for testing tactile sensor force resolution. The robot arm and stage setup are capable of precisely applying measured force onto a target device with controlled contact spatial location and orientation.}
        \label{fig:indenter}
    \end{minipage}
    \hspace{0.01\textwidth} 
    \begin{minipage}[b]{0.66\textwidth}
        \centering
        \includegraphics[width=\linewidth]{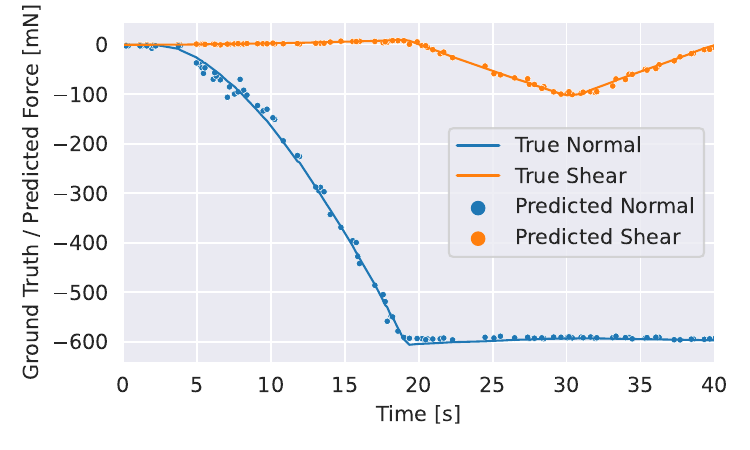} 
        \caption{Example of data collection and prediction. The solid line shows ground-truth shear and normal force trajectory during one indentation. The dotted scatter plot shows model-predicted shear and normal force.}
        \label{fig:contact-then-shear}
    \end{minipage}
\end{figure}

\subsection{Thin-film manufacturing and fabrication}
We manufacture the fingertip molds from $6061$ aluminum and finish them with a machine polishing pass with a \SI{3}{\milli\metre} diameter tool and a \SI{50}{\micro\metre} step-over. 
The molds are then prepared for gel casting through a silanization process in a desiccator with \SI{50}{\micro\litre} silane under vacuum for \SI{30}{\minute}. 
Following this, the gel material is prepared using a $1:1$ ratio of Smooth-On Solaris and combined in a speed mixer for \SI{3}{\minute} under vacuum to release any captured air in the sample. 
The gel material is then cast into the mold and allowed to cure at \SI{23}{\celsius} for \SI{12}{\hour}. 
Once cured, the gel fingertip is removed from the mold using tweezers for transfer to a glass slide. 
Here are the steps for preparing the thin film metallic reflective layer on the gel fingertip through silver plating.
First, a glucose solution is prepared by dissolving \SI{2.035}{\gram} glucose in \SI{160}{\milli\litre} \ce{H2O} and then adding \SI{0.224}{\gram} \ce{KOH}. This is set aside and the \ce{AgNO3} solution is prepared by dissolving \SI{1.02}{\gram} \ce{AgNO3} in \SI{120}{\milli\litre} \ce{H2O}, and then adding \SI{1.2}{\gram} \ce{NH3} $25\%$. The plating solution, which is used to silver coat the gel fingertip, is then prepared by mixing two parts glucose solution, total \SI{80}{\milli\litre}, to 1 part \ce{AgNO3} solution, total \SI{40}{\milli\litre}. 
The silvering solution is then set to stir gently. Prior to silver coating, the gel fingertip is cleaned using oxygen plasma for \SI{3}{\minute}. The gel fingertip is then activated in a solution of \SI{6.181}{\gram} \ce{SnCl2} in \SI{98}{\milli\litre} \ce{H2O} for \SI{10}{\second}. Once the gel fingertip is activated, it is suspended in the silvering solution for a total of \SI{3}{\minute}, then rinsed with \ce{H2O} and air dried. This process creates a silvered reflective layer with \SI{6}{\micro\metre} thickness. For robotics applications and for increasing resilience against the intrusion of ambient light, we coat the silvered layer in a white or black layer. This layer is produced by using Smooth-On Ecoflex $0010$, a mixing ratio of part A to B of $1:1$, and then adding $3~\%$ Smooth-On Silc Pig to part A of Ecoflex. Part B of Ecoflex is then mixed in by weight according to the mixing ratio specified previously and then mixed in the speed mixer for \SI{3}{\minute} under vacuum. The silvered gel fingertip is then dipped into the Ecoflex pigment and set to cure for \SI{6}{\hour}.

\begin{figure}[!t]
    \centering
    \begin{minipage}[b]{0.5\textwidth}
        \centering
        \subfloat[Max Normal Force Only $(t=20)$]{
            \includegraphics[width=0.45\textwidth]{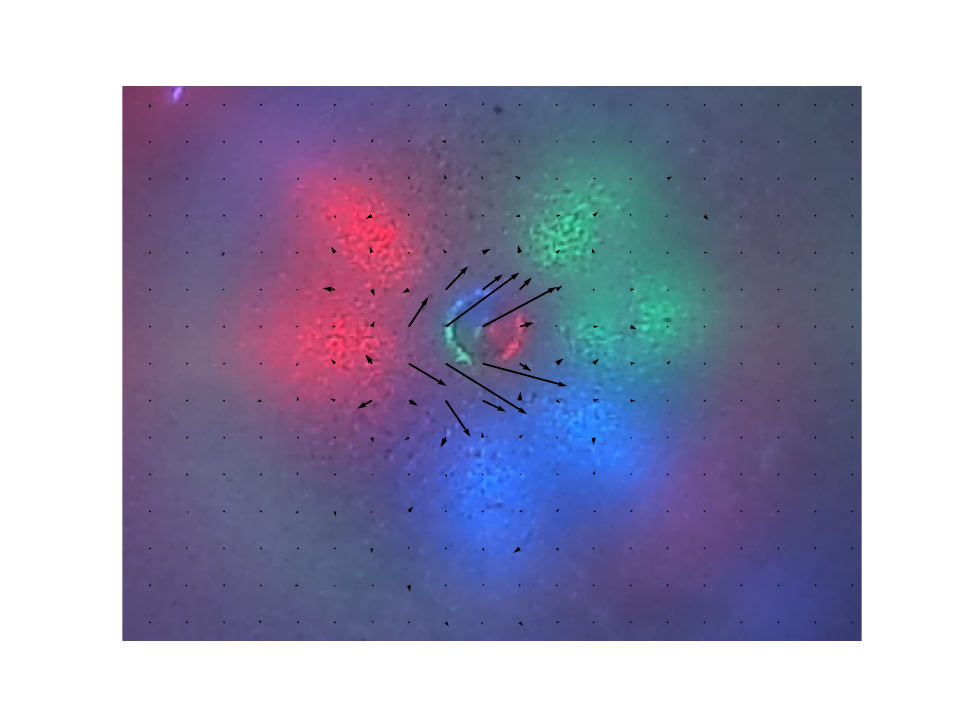}
        }
        \subfloat[Max Normal and Max Shear Force $(t=30)$]{
            \includegraphics[width=0.45\textwidth]{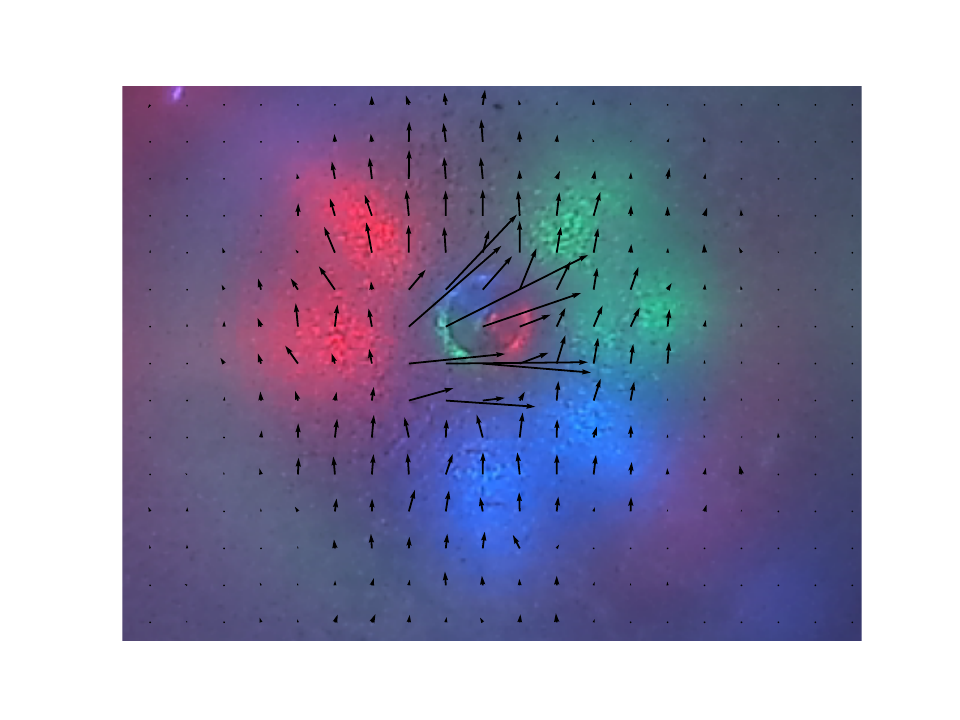}
        }
        \caption{\sensor{} image snapshots from shear force data collection in two key moments. The timestamp $t$ corresponds to the trajectory in Figure \ref{fig:contact-then-shear}. The overlay arrow field shows the optical flow w.r.t image without any force applied.}
    \label{fig:flow}
    \end{minipage}
    \hspace{0.01\textwidth} 
    \begin{minipage}[b]{0.45\textwidth}
    \centering
    \includegraphics[width=\linewidth]{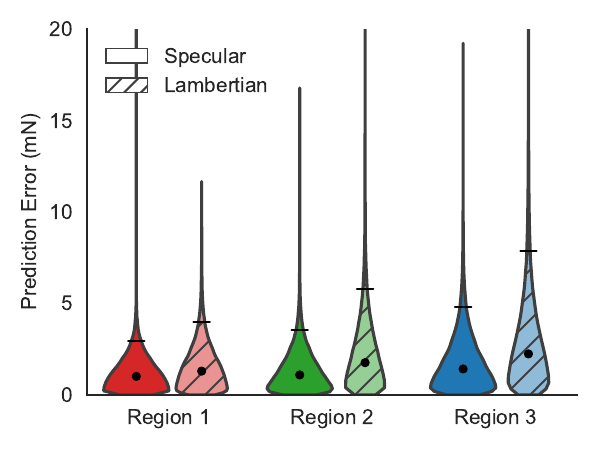}
    \caption{\sensor{} normal force prediction error distribution by surface types and regions. The dots and error bar show the median and 95-percentile of the error respectively.}
    \label{fig:force-scattering-comparison}
    \end{minipage}
\end{figure}

\subsection{Volume illumination simulation and evaluation}
Common vision-based tactile sensors make use of a static illumination configuration \citep{Gomes2020GelTip},\citep{Sun2022soft},\citep{Lambeta2020DIGIT} whereas sensors like Digger Finger \citep{patel2020digger} use a single light color with colored acrylic to simulate multiple colors. Static illumination is not ideal for promoting a modular system. Instead, the illumination system should adapt to the needs of extracting information from the touch surface. Previous work based on GelSight-style tactile sensors, \citep{Lambeta2020DIGIT}, \cite{Taylor2022gelslim}, \citep{Donlon2018gelslim}, \citep{do2022densetact},used a gel coated with a Lambertian scattering layer, in which volume illumination produces an image by means of scattering light off the surface and into the vision system. In the case of the monolithic hemispherical gel dome used in Digit 360, we determined that Lambertian scattering is not ideal for producing and optimizing for force and spatial sensitivity. Additionally, we introduced a dynamic illumination system that provides volume illumination with configurable wavelength, intensity, and positioning. The illumination system consists of 8 fully controllable RGB LEDs that emit Lambertian diffuse light, equally spaced around a circle of radius 9 mm.

\subsection{Internal surface scattering characterization}
Our system’s fingertip gel comprises three components, as seen in Figure \ref{fig:sensor-regions-lens}. The outer surface of the base gel has a reflective silver thin-film coating, which is coated with a protective colored diffusive material. To produce an image, the two layers provide scattering of internally incident light from surface interactions to the vision system. We placed the illuminating light-emitting diodes in optical contact with the gel, using an over-molding process. The fingertip gels are initially manufactured with a smooth and polished surface, and through mold texturing, we can control and determine how light is scattered at the interface.

Using a Gaussian scatter distribution, we modeled a range of scatter. Our scattering parameter, $\sigma$, was chosen to achieve the half-width-half-max angles, $\alpha$, of the bi-directional scatter distribution (BSDF) function at normal incidence from $\alpha$ = 1$\degree$ to 25$\degree$, along with a Lambertian scattering model. In Figure~\ref{fig:max-force}, we show that a surface texture that minimizes scattering produces little background illumination. Neither does it produce large shadows created by surface indentations. Additionally, minimizing scatter produces specular glint reflections, fails to illuminate all indentations equally, and saturates the vision system.

With a fully Lambertian scattering model, the hemispherical surface of the gel fingertip acts as an integrating sphere. While Lambertian scattering provides uniform background illumination, the high scattering illumination from nearby interactions reduces the overall indentation contrast. We optimized for high image contrast while maintaining uniform background illumination, better image impressions produced by gel indentations, and minimizing the number of glints that would saturate the image sensor. 

Two non-uniformity metrics were evaluated over the fingertip hemispherical surface, Std/Mean and (Max - Min)/Mean. We demonstrated that low scatter yields images with a significant variation in the image signal, requiring the camera to handle a high dynamic range. If the image is allowed to saturate, the visual signal resolution of region variations due to the indentations will be lost. Thus, in these cases, the stray light is more likely to cause objectionable artifacts. The contrast in the image caused by spherical indentations is in certain areas high, due to bright glint reflections; but regions with large gradients in the background may make the indentations hard to detect. High scatter gives images with low variation in image signal, and no areas are lost due to saturation. In the image caused by spherical indentations, the contrast is low in certain areas, but the uniform background makes these easier to detect. 

We define a contrast-to-noise (CNR) metric, and studied three regions of interest on the hemispherical surface for background uniformity noise and indentation contrast. Plotting the calculated CNR across the hemispherical surface for the different scatter angles, we saw that the CNR is generally higher for less scatter, but CNR is more uniform across the FOV for more scatter. Therefore, we determined that for a hemispherical fingertip surface, the desired texture scattering profile is constrained between half-width-half-max angles of 20$\degree$ to 25$\degree$.

\begin{figure}[t]   
    \centering
    \includegraphics[width=0.5\linewidth]{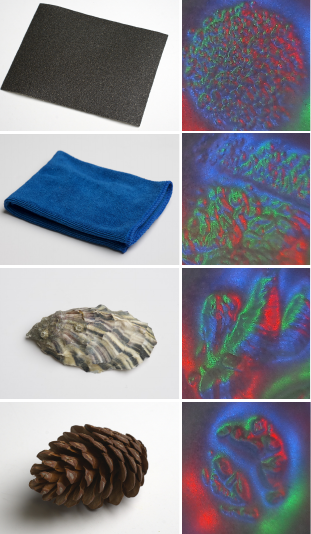} 
    \caption{We show a typical capture common amongst vision-based touch sensors using a red, blue, and green illumination scheme. Objects shown from top to bottom: sandpaper, cloth, oyster, and a tree pinecone.}
    \label{fig:output-rgb}
\end{figure}

\subsection{Edge AI at the fingertip}
Our system provides communications with the host device over the USB 3.0 standard interface. Three separate streams are provided for data transfer, supporting video, audio, and multi-modal data. These streams collectively output at a maximum rate of 148 MBps and below, depending on the configuration sent to the device. Currently, tactile sensors are used in open-loop control, providing information to the host device for processing and additional actions to manipulators. We propose adding edge AI at the fingertip for three reasons: First, to create a latent representation of the data and reduce the overall bandwidth sent to the host device. Second, to enable fast local decisions for transmitting actions to manipulators. Third, we aim to improve the overall latency of the system while reducing variance in jitter, which is the change in latency. We modeled the tactile fingertip system with a manipulator, where both systems were connected to the host device in a star configuration. This resulted in decisions — and actions resulting from tactile information — being processed through the host device and disseminated to the manipulators. As we move toward more data-intensive designs, capturing data from multiple fingers at once, this arrangement may result in unstable control schemes where information and action latency cannot be guaranteed. To accommodate and expand the terrain of tactile sensing research, we integrated the Greenwaves Technologies GAP9 neural network accelerator, a 9-core RISC-V compute cluster with AI acceleration, for on-device processing of selected data streams.

\begin{table}[t]
\centering
\setlength{\tabcolsep}{3pt}
\renewcommand{\arraystretch}{1.25}
\begin{tabular}{lllll}
Surface & Region 1 [mN] & Region 2 [mN] & Region 3 [mN] & Mean [mN]\\
\hline
Specular  & 1.01 & 1.09 & 1.41  & 1.17 \\
Lambertian & 1.30 & 1.77 & 2.24  & 1.77 \\
\end{tabular}
\caption{\sensor{} normal force prediction error (median) by surface types and regions.}
\label{tbl:force-by-scattering}
\end{table}

\subsection{Timing characterization of Reflex Arc with Edge AI }
Following high-level abstractions of the human reflex arc \citep{Dewey1896reflex}, we developed a fast reflex-like control loop using Edge AI for local processing. The current paradigm of transferring the sensory input to a central control computer for processing and then sending back the control signals requires high bandwidth while introducing communication latency. In contrast, our paradigm is to process the sensory input inside the fingertip locally using Edge AI. This allows drastic reductions of the required bandwidth while significantly minimizing communication latency and — most importantly— minimizing jitter. We performed an experimental comparison of these two paradigms, shown in Table~\ref{tbl:force-by-scattering}, by measuring the end-to-end latency of the systems using a PCI-e-based precision time measurement tool. We evaluated this experiment on a Linux machine with an Intel i9-11900K, 64GB memory, and an NVIDIA Quadro RTX5000 GPU. First, to ensure granularity in the measurements, each section of the system was isolated, and samples were collected in repeated trials. Second, we verified these results by subjecting the entire system to repeated measurements and comparing the timing results to the sum of the isolated components. This determined the areas that produce deterministic timing results, as well as highlighting the areas that sustained increased latency and jitter. Furthermore, these results indicated areas of performance improvements and design for future tactile sensors. The results for the entire control loop show how the Edge AI paradigm results in a reduction of latency from 4 ms to 1 ms with a desirable smaller variance. It is also noteworthy that, in principle, appropriate Edge-AI processing could be extended further to exploit the sequential nature of the camera FIFO memory to parallelize the data capture with the processing, thus yielding even lower latency. In this case, instead of processing the entire image, select horizontal lines would be sent for processing in the configured region of interest. This is applicable when touch interactions are most likely to appear in certain regions on our omnidirectional fingertip. Our system supports this region-of-interest data output selection for increased resolution and image-capture frequency.

\begin{figure}[!t]
    \centering
    \includegraphics[width=\linewidth]{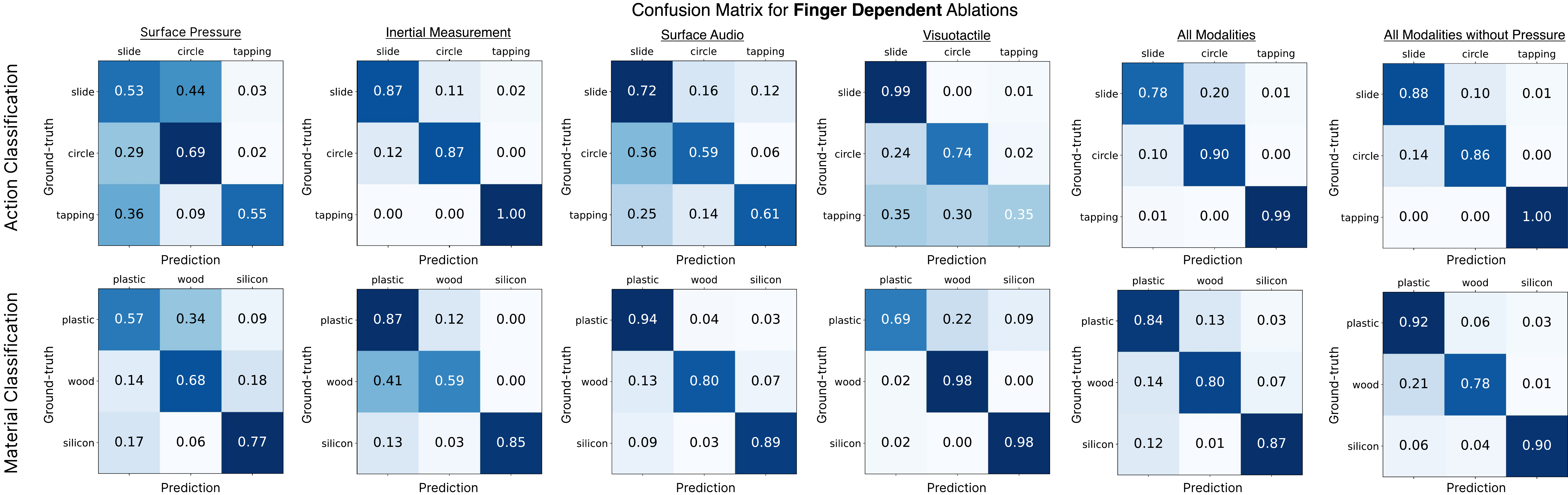}
    \caption{Confusion matrix for the finger-dependent ablation multi-modal evaluation.}
    \label{fig:cls_cfs_matrix_finger_dependent}
\end{figure}

\begin{figure}[!ht]
    \centering
    \includegraphics[width=\linewidth]{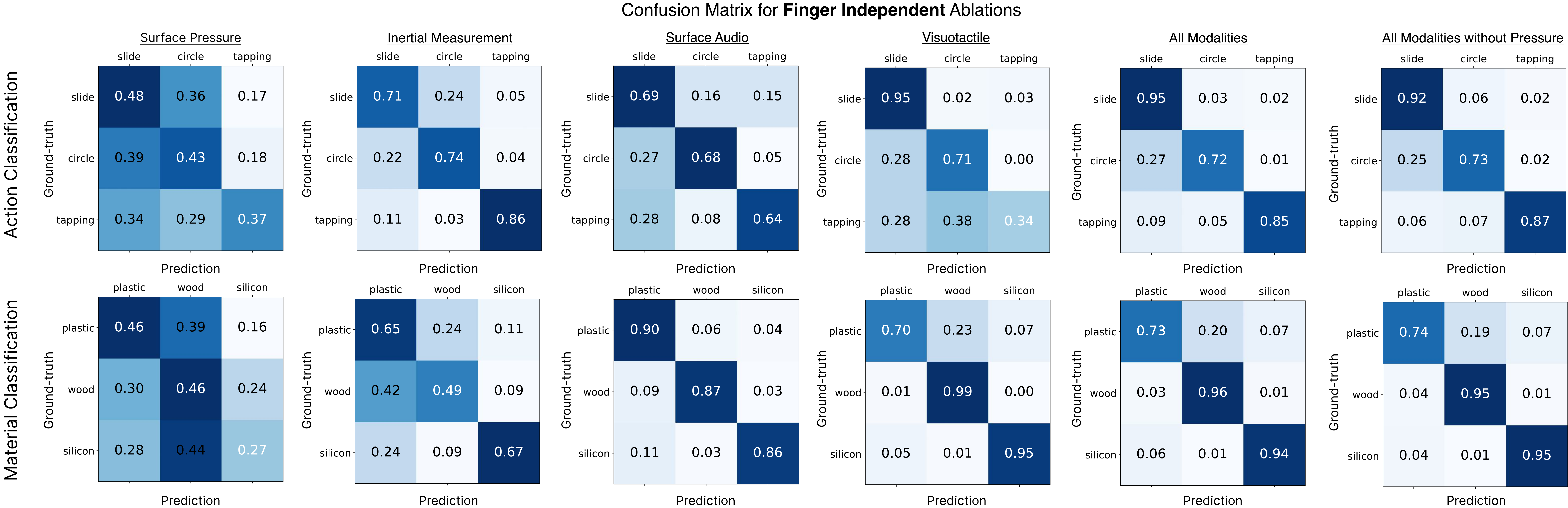}
    \caption{Confusion matrix for the finger-independent ablation multi-modal evaluation.}
    \label{fig:cls_cfs_matrix_finger_independent}
\end{figure}

We established two pipelines for data transfer and processing: on-device pipelines and host pipelines, shown in Figure~\ref{fig:mm-water}. Additionally, a hybrid mode supports transferring data to the host and processing it on the device. We observed that this affects the vision system because it involves substantially larger amounts of data than does the multi-modal data system.
The limiting parameter for dynamic manipulations in a task is the latency between input data, processing, and action commands. To elicit rapid responses to changes in the environment that are detected through changes in grasp stability, we must move toward faster processing and minimizing latency between input and action.

We studied the effects of the vision system because this is the most common modality used in touch sensors, with impacts on overall system latency for host and on-device configurations. The constraining factor on system latency between real-world input and data-processing input is the capture rate of the vision system. This is limited by the frames per second rate, which imposes a delay of 1/fps, and the internal processing of the image-signal processor. For this reason, we designed our system to incorporate a CMOS sensor with 240 fps and a pixel size of 1.1~um, which yields a shorter delay (d) of 4.17 ms as opposed to prior sensors such as Digit \citep{Lambeta2020DIGIT}, which operate at 60 Hz and thus have delays, d $\ge$ 16.7 ms.

We evaluated the inference latency with two deep neural networks, MLP and MobileNetV2 \citep{sandler2018mobilenetv2}, for two scenarios:  on-device and host inference. The two most significant sources of latency arise from the tactile data transfer from device to host and of action data from host to robotic end effector. We determined that within the available headroom between the differences in pipeline latency and Tlatency, an upper limit of Tlatency $\le$ 2.463 ms was established. With an MLP-based network, we increased the layer depth and observed the latency cost for both scenarios. As shown in Figure~\ref{fig:sensor-regions-lens}, it becomes apparent that using the on-device accelerator without enabling the hardware engine quickly exceeds Tlatency at 10-layer depth. However, enabling hardware acceleration allows us to use MLP models with 60 layers.
Observing a more suitable use case for the tactile research domain, we deployed a MobileNetV2 model and determined the total system pipeline latency. We show in Figure~\ref{fig:edgeai-data-timing-mlp} that using a MobileNetV2 architecture with an input size of 64 × 64 is beneficial for reducing Tlatency and applicable to common low-level touch tasks, such as touch detection \citep{lambeta2021pytouch} and classification \citep{zhang2021target}. Furthermore, the Tlatency upper boundary is determined by the output data rate, size of data transfer, and the host system performance.

However, in practical robotics environments, the host system is running a plethora of control and processing applications, with the additional overhead of communication between other sensors and devices that introduce overhead to a Tlatency of 1.2 ms. Comparing this to Digit \citep{Lambeta2020DIGIT}, we observed an overhead of 4.7 ms. These differences are attributed to the frame rate of Digit 360, 240fps, and data transfer over USB 3.0, whereas Digit is limited to 60fps using USB 2.0. While the performance gains of an on-device AI accelerator for a low-powered device such as an artificial fingertip may not be significantly fruitful today, regarding stark latency differences, we believe on-device AI accelerators will only improve in the future. Our system enables a first look into exploring this early stage of tactile on-device inference with low latency control. The primary motivation was driven by the ability to have reflex-like control of the device to which the system is connected and provide to the host abstractions of lower-level touch signals. An example would be training an on-device model to regress force from multi-modal data to introduce touch and manipulation force limits to objects. Another example would be using the on-device AI capabilities to recognize slip and, with low latency, provide actions to the robotic end effector to reconfigure grasp.

\subsection{Dataset collection for multimodal action and material classification}
To collect a multimodal action and material dataset, we construct a setup using a Franka Emika robotic arm, a Wonik Allegro robotic hand, and four Digit 360 sensors attached to the distal joint of each finger, serving as the fingertip. A set of motion trajectories is programmed to execute various actions that translate, rotate, and perturrate the robotic hand. Data is streamed from the devices to a host computer in which parameters and signals from the arm, hand, and fingertips are saved. We collect the following trajectories: four-finger grasp, slide, stir, tap, and translation and rotation perturbations. The total dataset size collected results in $\approx$ 1.1 million 1.33~s window samples across four fingers and 12 modalities, of which $\approx$ 624k samples are for action and material classification. The remaining dataset comprises multimodal capture for supporting experiments listed in the figures. To determine the window sample size of 1.33~s, we shift the window size length in order to include more than one periodic signal episode that reflect different actions across the modalities. The original raw collected dataset is then resampled such that all modalities align with this input frequency.

\subsection{Multimodal action and material classification analysis}
Since each modality is represented as a spatial, temporal, or a combination of the two, we preprocess the dataset items based on their respective modality. In the case of visuotactile inputs, represented as a 3-channel RGB image, we center crop and downsample the hyperfisheye such that the result is a slightly underfilled 160x160 pixel image. For the surface audio signals, we take each window sample and compute the MEL-spectrogram with the following parameters: $n_{fft}=2048$ (number of FFT bins), and $n_{overlap}=1024$ (overlap stride between each sample), and then scale the resulting RGB spectrogram to 64x64 pixels. The inertial measurement signals are concatenated for each axis and each window sample. For the surface pressure modality, we preprocess the raw signal with two series filters, first a high pass filter with $f_{c}=0.95$ Hz (frequency cutoff), and then a low pass of $f_{c}=50$ Hz to extract saliant features particular to the perturbations the fingertips are subjected to during dataset collection, for example, sensitive to rapid fingertip adjustments, but filtering out static forces experienced by the grasp of each finger on the object. The other modalities are treated as is without any pre-processing.
We construct a model comprising multiple multimodal inputs for RGB and temporal encoders. For the RGB encoder, we use a modified, non-pre-trained ResNet-18 model with the following modifications: batch norm layers changed to group norm layers for training stability, and for the surface audio input, we change the fully connected layer to identity to allow for propagation from and to other blocks. We construct a sequential two-hidden layer deep network for the temporal layers. For training all the modalities, we first encode each modality, and the output of the modality encoder is concatenated to a final MLP, which provides a multi-head output in classifying actions and materials. 
Finally, we deploy model training with a maximum of 200 epochs with a grid search optimization of hyperparameters for the Adam optimizer learning rate. We present the results in~\ref{tab:multimodal-action-material-cls}, and show the confusion matrix for both finger dependent~\ref{fig:cls_cfs_matrix_finger_dependent} and finger independent~\ref{fig:cls_cfs_matrix_finger_independent} ablations.

\subsection{Dataset collection for normal and shear-force analysis}
We designed a controllable robot indenter capable of applying a high-precision measured 3-axis force to any spatial position of the sensor. As shown in Figure~\ref{fig:edgeai-data-pipeline}, a tactile sensor is mounted on the robot arm (Mecademic Robotics Meca500) to orient the desired test surface down against a probe with a precision of 5 µm. The probe with a hemisphere tip of 4 mm diameter is mounted on a force sensor (FUTEK QMA147 3-axis or ME-Systems KD34s single axis) measuring the ground-truth contact force with 1 mN accuracy. The probe and force sensor assembly are then mounted on a hexapod (Newport HXP50), which can be precisely controlled to translate with 0.1 µm and rotate with 0.05\degree increment. Due to the rotational symmetry of the artificial fingertip, we break down the full-surface force characterization into three representative approximately planar regions, as shown in Figure~\ref{fig:gel-layers}. We repeat the collection process for each region similarly for normal and shear forces. 

We start with normal force collection, and for high precision, we use the single-axis force sensor that can measure up to 250 mN. The robotic indenter spatially samples 0.5mm-spaced grid points for each region on the tangential plane. The probe moves perpendicular to the plane for each point, pressing into the sensor until the normal force reaches 200 mN. During the contact between the probe and gel (defined as normal force, Fnorm $>$ 0.2 mN), both sensor images and measured normal force are collected synchronously. Empirically, we collect about 550 image-force pairs per spatial point. For a 7~mm×~6~mm region, we obtain approximately 12,000 points. This point data is randomly split into training (70\percent) and testing set (30\percent).

For shear-force data collection, we selected a 3-axis force sensor to measure normal and shear forces simultaneously. We needed to apply sufficient friction while varying shear force. Figure~\ref{fig:contact-then-shear} shows how each shear-force indentation trajectory was controlled. First, the probe was moved perpendicular to the contact surface to apply up to 600 mN of normal force. Next, the probe was moved tangential to the surface, loading shear force up to 100 mN. Finally, the probe was returned to the previous location, unloading shear force. If non-zero residual shear force remained after unloading, slip might have occurred, in which case the data is discarded.

\subsection{Image-to-force regression model analysis}
Contact-force prediction on vision-based tactile sensors such as our system can be achieved using an image-to-force regression model. The model needs to be calibrated from reference data. Once calibrated, evaluating the sensor model as a system for force-sensing performance on a testing dataset is possible. In this section, we share how we collected the dataset — training and evaluating the model to benchmark normal and shear-force-sensing performance. We used a modified ResNet50 \citep{he2016deep} deep neural network for the image-to-force regression model. The network was initially designed to take an input image of 224×224×3 and output 1024-way object-classification probabilities. We replaced the classification head with a scalar-output linear layer predicting the force. We used mean-square error as our loss, then optimized with Adam with an initial learning rate search. The raw images from the sensors are 640×480, which were down-scaled to 224×224 with 20-pixel spatial jitter to improve spatial invariance. We pooled training data from all three regions to train a single model and obtain the prediction performance (median error) breakdown by regions as described in Figure \ref{fig:fig_1}A.

\begin{table}[]
\centering
\setlength{\tabcolsep}{3pt}
\renewcommand{\arraystretch}{1.25}
\resizebox{\textwidth}{!}{%
\begin{tabular}{l|lll|lll}
\hline
                     & Independent &          &         & Dependent &          &         \\ \hline
Modality             & Action      & Material & Average & Action    & Material & Both \\ \hline
Surface Pressure     & $43.30\pm1.50$   & $39.52\pm0.68$  & $41.41\pm0.76$     & $60.37\pm2.20$   & $66.43\pm4.17$     & $63.40\pm2.73$      \\
Surface Audio        & $67.54\pm0.61$   & $87.84\pm0.62$  & $77.69\pm0.26$     & $63.82\pm3.56$   & $88.16\pm2.56$     & $75.99\pm3.82$     \\
Inertial Measurement & $75.42\pm0.93$   & $61.15\pm0.65$  & $68.29\pm0.51$     & $90.04\pm1.31$   & $78.31\pm1.71$     & $84.17\pm1.09$     \\
Visuotactile         & $71.43\pm0.86$   & $86.60\pm2.19$  & $79.01\pm1.03$     & $74.16\pm4.29$   & $87.02\pm4.31$     & $80.59\pm3.82$     \\ \hline
All                  & $82.68\pm1.92$   & $86.41\pm2.92$  & $84.55\pm0.85$     & $87.90\pm1.96$   & $83.68\pm7.83$     & $85.79\pm3.79$      \\ \hline
\end{tabular}%
}
\caption{We compare two ablations with varying inputs to the multi-modal deep neural network, finger-dependent and independent inputs. Here, we show the performance across each modality and all modalities for actions (slide, tap, stir) and materials (wood, plastic, silicone).}
\label{tab:multimodal-action-material-cls}
\end{table}

\subsection{Force resolution factors and effects on shear force detection}
Table~\ref{tab:multimodal-action-material-cls} and Figure~\ref{fig:flow} show additional normal-force resolution performance with two kinds of gel surface finish: specular and Lambertian. We consistently saw that Lambertian surface scattering, typically considered preferable for vision-based tactile sensors, is, in fact, outperformed by its specular counterpart. This may come from the enhancement of surface texture contrast due to specular reflection, which helps the imaging system track gel deformation. Figure~\ref{fig:indenter} shows the center-crop of our system’s image, where the texture contrast is more evident in the region with strong specular reflections from the LEDs. We obtained clear optical flow (shown in the figure by arrows) within these texture-rich regions, corresponding to fingertip deformation caused by shear and normal force applied by the probe.

Previously, a general view held that some tracking pattern (dots, for example) was required for shear force measurement. However, the optical flow result reported above suggests that this requirement can be relaxed due to the increasing resolution and quality of images. Such advancements facilitate using the natural fingertip surface texture to observe gel deformation and, in turn, to perform shear force estimation.

\begin{table*}[!t]
\resizebox{\textwidth}{!}{%
\begin{tabular}{l|llllllll}
\multicolumn{1}{c|}{\textbf{Sensor}} &
  \multicolumn{1}{c}{\textbf{Technology}} &
  \multicolumn{1}{c}{\textbf{Modalities}} &
  \multicolumn{1}{c}{\textbf{\begin{tabular}[c]{@{}c@{}}Sample Rate \\ {[}\SI{}{\hertz}{]}\end{tabular}}} &
  \multicolumn{1}{c}{\textbf{\begin{tabular}[c]{@{}c@{}}Area \\ {[}$mm^2${]}\end{tabular}}} &
  \multicolumn{1}{c}{\textbf{\begin{tabular}[c]{@{}c@{}}Spatial\\ Resolution \\ {[}$m${]}\end{tabular}}} &
  \multicolumn{1}{c}{\textbf{\begin{tabular}[c]{@{}c@{}}Normal \\ Force\\ Resolution \\ {[}$N${]}\end{tabular}}} &
  \multicolumn{1}{c}{\textbf{\begin{tabular}[c]{@{}c@{}}Shear \\ Force\\ Resolution \\ {[}$N${]}\end{tabular}}} \\ \hline
Human finger &
   &
  \begin{tabular}[c]{@{}l@{}}Contact location, pressure, \\ texture, temperature\end{tabular} &
  1000 &
  -- &
  1 &
  0.06 &
  -- &
   \\
BioTac &
  Fluid, hydrophone &
  Contact force, location, pressure &
  100 &
  484 &
  1.4 &
  0.26 &
  0.48 &
   \\
GelSight &
  Visual, markers &
  RGB image &
  30 &
  252 &
  0.03 &
  0.66 &
  0.17 &
   \\
GelSlim &
  Visual, markers &
  RGB image &
  60 &
  -- &
  0.03 &
  0.32 &
  0.22 &
   \\
Insight &
  Visual &
  RGB image &
  11 &
  4800 &
  0.4 &
  0.03 &
  0.03 &
   \\
ReSkin &
  Magnetic &
  Contact force, location, pressure &
  400 &
  400 &
  2.5 &
  0.2 &
  -- &
   \\
Epstein et al. &
  Pressure &
  Contact force, location, pressure &
  200 &
  184 &
  1.11 &
  0.65 &
  -- &
   \\
SaLoutos et al. &
  Pressure, time-of-flight &
  Contact force, location, proximity &
  200 &
  406 &
  1.31 &
  1.58 &
  -- &
   \\ 
   OmniTact &
  Visual, omnidirectional, multi-camera &
  RGB image &
  30 &
  -- &
  0.4 &
  -- &
  -- &
   \\ 
   Digit &
  Visual &
  RGB image &
  60 &
  304 &
  0.150 &
  0.006 &
  0.012 &
   \\ \hline
Digit 360 &
  \begin{tabular}[c]{@{}l@{}}Visual, pressure, \\temperature, audio, \\ EdgeAI\end{tabular} &
  \begin{tabular}[c]{@{}l@{}}RGB image, contact location, \\ pressure, contact force, \\ contact audio texture\end{tabular} &
  10,000 &
  2340 &
  0.007 &
  0.0010 &
  0.0013
\end{tabular}%
}
\caption{Characterization and performance of commonly used sensors in touch perception for robotic manipulation.}
\label{tab:sensors-comparison}
\label{tab:alternatives}
\end{table*}

\subsection{Environmental and local gas sensing}
We establish a modality within the artificial fingertip to determine the object state and obtain clues on object classification. We identify two key performance metrics for fingertip gas sensing: accuracy and signal acquisition time. We observe six different materials, from liquid to solid, commonly found in a household environment. These materials are coffee powder, liquid coffee, a nondescript rubber material, cheese, and a spread of soap and butter on a surface. All the materials were sampled at room temperature with a Franka robotic arm and Digit 360 approaching the samples to near contact, within 1cm, for a duration of 90s. We record multi-modal data and isolate the humidity, temperature, pressure, and gas oxidation resistance data points at the maximum output frequency for each sampling modality. Over 100 approaches to each material are collected during a 3-hour sampling period. Between each approach, we sample air from the local environment. The raw data with the modalities of interest listed above are provided as inputs to a multi-layer perceptron network with a single 64-node hidden layer. We train the network with cross-entropy loss using an Adam optimizer with a learning rate of 0.1. We further ablate this study to show that the final accuracy of the model is not sensitive with respect to the size of hidden layers or learning rate. We show that through these six materials, a classification accuracy of 91\percent. Furthermore, we show the signal acquisition time to reach 66\percent~accuracy.

\subsection{Sensor comparison to alternatives}
In Table~\ref{tab:alternatives}, we compare the performance of our system to some of the most commonly used tactile sensors in touch manipulation research at the time of this writing.

\section*{Acknowledgments}
We thank Kirill Belyayev and Velentium LLC for help with electronics CAD layout; Jim Hatlo for leading the editorial styling and technical writing of the manuscript; the Leibniz Institute for New Materials for gel process engineering and characterization of the elastomers; GreenWaves Technologies SAS for providing their GAP9 processor; Bosch for providing samples of their smart-hub platform and firmware support; Greenlight Optics LLC for manufacturing the lens prototypes; and Percipio Robotics for characterization of the sensor. 
R.Ca. discloses support for the research of this work from the German Research Foundation (DFG, Deutsche Forschungsgemeinschaft) as part of Germany’s Excellence Strategy – EXC 2050/1 – Project ID 390696704 – Cluster of Excellence “Centre for Tactile Internet with Human-in-the-Loop” (CeTI) of Technische Universität Dresden, and by Bundesministerium für Bildung und Forschung (BMBF) and German Academic Exchange Service (DAAD) in project 57616814 (\href{https://secai.org/}{SECAI}, \href{https://secai.org/}{School of Embedded and Composite AI}).

\section*{Declarations}

\textbf{Competing interests.}
Meta Platforms Technologies, LLC filed a provisional patent application 63/383,069, titled ``Digitizing Touch with an Artificial Fingertip'' that describes one or more concepts discussed in this paper.

\textbf{Authors' contributions.}
R.Ca. and M.L. conceptualized the sensor design.
N.B., G.K., and M.L. designed the mechanics of the sensor.
M.L. designed the electronics of the sensor.
R.Ch. and J.K. supported the manufacturing and assembly of the electronics.
K.S., R.M., T.C., A.So., K.J., B.T., and Y.D. designed the sensor's optical system.
V.R.S., K.M., and N.S. investigated, designed, and manufactured the elastomer.
G.K., D.S., E.S., and N.T. manufactured the mechanics of the sensor.
R.Ca., M.L., T.W., A.Se. and J.M. conceptualized the experiments.
T.W. and M.L. designed and conducted experiments on edge AI.
A.Se., T.W., and M.L. and H.Q. designed and conducted experiments on sensor characterization.
H.Q. designed the architecture for the multimodal deep neural network.
E.S. and N.T. helped with the instrumentation for sensor characterization.
R.Ca., M.L., and T.W. drafted the manuscript and revised it.
R.Ca., M.L., and J.M. were responsible for the project administration.
R.Ch. and A.Se. organized and sourced 3rd party vendors for the project.

\clearpage
\newpage
\bibliographystyle{assets/plainnat}
\bibliography{paper}

\begin{thebibliography}{30}
\providecommand{\natexlab}[1]{#1}
\providecommand{\url}[1]{\texttt{#1}}
\expandafter\ifx\csname urlstyle\endcsname\relax
  \providecommand{\doi}[1]{doi: #1}\else
  \providecommand{\doi}{doi: \begingroup \urlstyle{rm}\Url}\fi

\bibitem[Abad and Ranasinghe(2020)]{Abad2020Visuotactile}
Alexander~C. Abad and Anuradha Ranasinghe.
\newblock Visuotactile sensors with emphasis on gelsight sensor: A review.
\newblock \emph{IEEE Sensors Journal}, 2020.

\bibitem[Ardiel and Rankin(2010)]{Ardiel2010importance}
Evan~L Ardiel and Catharine~H Rankin.
\newblock The importance of touch in development.
\newblock \emph{Paediatrics \& child health}, 2010.

\bibitem[Calandra et~al.(2017)Calandra, Owens, Upadhyaya, Yuan, Lin, Adelson, and Levine]{Calandra2017Feeling}
Roberto Calandra, Andrew Owens, Manu Upadhyaya, Wenzhen Yuan, Justin Lin, Edward~H. Adelson, and Sergey Levine.
\newblock The feeling of success: Does touch sensing help predict grasp outcomes?
\newblock In \emph{Conference on Robot Learning (CoRL)}, 2017.

\bibitem[Choi and Tahara(2020)]{Choi2020Dexterous}
Seung-hyun Choi and Kenji Tahara.
\newblock Dexterous object manipulation by a multi-fingered robotic hand with visual-tactile fingertip sensors.
\newblock \emph{ROBOMECH Journal}, 2020.

\bibitem[Dahiya et~al.(2009)Dahiya, Metta, Valle, and Sandini]{Dahiya2009Tactile}
Ravinder~S Dahiya, Giorgio Metta, Maurizio Valle, and Giulio Sandini.
\newblock Tactile sensing -- from humans to humanoids.
\newblock \emph{IEEE Transactions on Robotics (T-RO)}, 2009.

\bibitem[Dewey(1896)]{Dewey1896reflex}
John Dewey.
\newblock The reflex arc concept in psychology.
\newblock \emph{Psychological review}, 1896.

\bibitem[Do and Kennedy(2022)]{do2022densetact}
Won~Kyung Do and Monroe Kennedy.
\newblock Densetact: Optical tactile sensor for dense shape reconstruction.
\newblock In \emph{International Conference on Robotics and Automation (ICRA)}, 2022.

\bibitem[Donlon et~al.(2018)Donlon, Dong, Liu, Li, Adelson, and Rodriguez]{Donlon2018gelslim}
Elliott Donlon, Siyuan Dong, Melody Liu, Jianhua Li, Edward Adelson, and Alberto Rodriguez.
\newblock Gelslim: A high-resolution, compact, robust, and calibrated tactile-sensing finger.
\newblock In \emph{International Conference on Intelligent Robots and Systems (IROS)}, 2018.

\bibitem[Dunbar(2010)]{Dunbar2010social}
Robin~IM Dunbar.
\newblock The social role of touch in humans and primates: behavioural function and neurobiological mechanisms.
\newblock \emph{Neuroscience \& Biobehavioral Reviews}, 2010.

\bibitem[Gomes et~al.(2020{\natexlab{a}})Gomes, Lin, and Luo]{Gomes2020}
Daniel~Fernandes Gomes, Zhonglin Lin, and Shan Luo.
\newblock Blocks world of touch: Exploiting the advantages of all-around finger sensing in robot grasping.
\newblock \emph{Frontiers in Robotics and AI}, 2020{\natexlab{a}}.

\bibitem[Gomes et~al.(2020{\natexlab{b}})Gomes, Lin, and Luo]{Gomes2020GelTip}
Daniel~Fernandes Gomes, Zhonglin Lin, and Shan Luo.
\newblock Geltip: A finger-shaped optical tactile sensor for robotic manipulation.
\newblock In \emph{International Conference on Intelligent Robots and Systems (IROS)}, 2020{\natexlab{b}}.

\bibitem[Handler and Ginty(2021)]{Handler2021mechanosensory}
Annie Handler and David~D Ginty.
\newblock The mechanosensory neurons of touch and their mechanisms of activation.
\newblock \emph{Nature Reviews Neuroscience}, 2021.

\bibitem[Hayward(2011)]{Hayward2011Is}
Vincent Hayward.
\newblock Is there a ‘plenhaptic’function?
\newblock \emph{Philosophical Transactions of the Royal Society B: Biological Sciences}, 2011.

\bibitem[He et~al.(2016)He, Zhang, Ren, and Sun]{he2016deep}
Kaiming He, Xiangyu Zhang, Shaoqing Ren, and Jian Sun.
\newblock Deep residual learning for image recognition.
\newblock In \emph{Conference on Computer Vision and Pattern Recognition (CVPR)}, 2016.

\bibitem[Johansson and Flanagan(2009)]{Johansson2009Coding}
Roland~S Johansson and J~Randall Flanagan.
\newblock Coding and use of tactile signals from the fingertips in object manipulation tasks.
\newblock \emph{Nature Reviews Neuroscience}, 2009.

\bibitem[Johansson and Vallbo(1979)]{Johansson1979Tactile}
Roland~S Johansson and Ake~B Vallbo.
\newblock Tactile sensibility in the human hand: relative and absolute densities of four types of mechanoreceptive units in glabrous skin.
\newblock \emph{The Journal of physiology}, 1979.

\bibitem[Johnson(2001)]{Johnson2001roles}
Kenneth~O Johnson.
\newblock The roles and functions of cutaneous mechanoreceptors.
\newblock \emph{Current Opinion in Neurobiology}, 2001.

\bibitem[Klatzky and Lederman(1992)]{klatzky1992stages}
Roberta~L Klatzky and Susan~J Lederman.
\newblock Stages of manual exploration in haptic object identification.
\newblock \emph{Perception \& psychophysics}, 1992.

\bibitem[Lambeta et~al.(2020)Lambeta, Chou, Tian, Yang, Maloon, Most, Stroud, Santos, Byagowi, Kammerer, Jayaraman, and Calandra]{Lambeta2020DIGIT}
Mike Lambeta, Po-Wei Chou, Stephen Tian, Brian Yang, Benjamin Maloon, Victoria~Rose Most, Dave Stroud, Raymond Santos, Ahmad Byagowi, Gregg Kammerer, Dinesh Jayaraman, and Roberto Calandra.
\newblock {DIGIT}: A novel design for a low-cost compact high-resolution tactile sensor with application to in-hand manipulation.
\newblock \emph{IEEE Robotics and Automation Letters (RA-L)}, 2020.

\bibitem[Lambeta et~al.(2021)Lambeta, Xu, Xu, Chou, Wang, Darrell, and Calandra]{lambeta2021pytouch}
Mike Lambeta, Huazhe Xu, Jingwei Xu, Po-Wei Chou, Shaoxiong Wang, Trevor Darrell, and Roberto Calandra.
\newblock Pytouch: A machine learning library for touch processing.
\newblock In \emph{International Conference on Robotics and Automation (ICRA)}, 2021.

\bibitem[Lederman and Klatzky(1987)]{lederman1987hand}
Susan~J Lederman and Roberta~L Klatzky.
\newblock Hand movements: A window into haptic object recognition.
\newblock \emph{Cognitive Psychology}, 1987.

\bibitem[Li et~al.(2020)Li, Kroemer, Su, Veiga, Kaboli, and Ritter]{Li2020review}
Qiang Li, Oliver Kroemer, Zhe Su, Filipe~Fernandes Veiga, Mohsen Kaboli, and Helge~Joachim Ritter.
\newblock A review of tactile information: Perception and action through touch.
\newblock \emph{IEEE Transactions on Robotics (T-RO)}, 2020.

\bibitem[Padmanabha et~al.(2020)Padmanabha, Ebert, Tian, Calandra, Finn, and Levine]{Padmanabha2020OmniTact}
Akhil Padmanabha, Frederik Ebert, Stephen Tian, Roberto Calandra, Chelsea Finn, and Sergey Levine.
\newblock Omnitact: A multi-directional high-resolution touch sensor.
\newblock In \emph{International Conference on Robotics and Automation (ICRA)}, 2020.

\bibitem[Patel et~al.(2020)Patel, Ouyang, Romero, and Adelson]{patel2020digger}
Radhen Patel, Rui Ouyang, Branden Romero, and Edward Adelson.
\newblock Digger finger: Gelsight tactile sensor for object identification inside granular media.
\newblock In \emph{International Symposium on Experimental Robotics}, 2020.

\bibitem[Romero et~al.(2020)Romero, Veiga, and Adelson]{Romero2020Soft}
Branden Romero, Filipe Veiga, and Edward Adelson.
\newblock Soft, round, high resolution tactile fingertip sensors for dexterous robotic manipulation.
\newblock In \emph{International Conference on Robotics and Automation (ICRA)}, 2020.

\bibitem[Sahli et~al.(2020)Sahli, Prot, Wang, M{\"u}ser, Piovar{\v{c}}i, Didyk, and Bennewitz]{sahli2020tactile}
Riad Sahli, Aubin Prot, Anle Wang, Martin~H M{\"u}ser, Michal Piovar{\v{c}}i, Piotr Didyk, and Roland Bennewitz.
\newblock Tactile perception of randomly rough surfaces.
\newblock \emph{Nature Scientific Reports}, 2020.

\bibitem[Sandler et~al.(2018)Sandler, Howard, Zhu, Zhmoginov, and Chen]{sandler2018mobilenetv2}
Mark Sandler, Andrew Howard, Menglong Zhu, Andrey Zhmoginov, and Liang-Chieh Chen.
\newblock Mobilenetv2: Inverted residuals and linear bottlenecks.
\newblock In \emph{Conference on Computer Vision and Pattern Recognition (CVPR)}, 2018.

\bibitem[Sun et~al.(2022)Sun, Kuchenbecker, and Martius]{Sun2022soft}
Huanbo Sun, Katherine~J Kuchenbecker, and Georg Martius.
\newblock A soft thumb-sized vision-based sensor with accurate all-round force perception.
\newblock \emph{Nature Machine Intelligence}, 2022.

\bibitem[Taylor et~al.(2022)Taylor, Dong, and Rodriguez]{Taylor2022gelslim}
Ian~H Taylor, Siyuan Dong, and Alberto Rodriguez.
\newblock Gelslim 3.0: High-resolution measurement of shape, force and slip in a compact tactile-sensing finger.
\newblock In \emph{International Conference on Robotics and Automation (ICRA)}, 2022.

\bibitem[Zhang et~al.(2021)Zhang, Li, Yang, Bai, Wang, Shen, Pu, and Song]{zhang2021target}
Xingxing Zhang, Shaobo Li, Jing Yang, Qiang Bai, Yang Wang, Mingming Shen, Ruiqiang Pu, and Qisong Song.
\newblock Target classification method of tactile perception data with deep learning.
\newblock \emph{Entropy}, 2021.

\end{thebibliography}

\end{document}